\documentclass{article}

\usepackage{arxiv}

\usepackage[utf8]{inputenc} % allow utf-8 input
\usepackage[T1]{fontenc}    % use 8-bit T1 fonts
\usepackage[arabic,english]{babel}
\usepackage{hyperref}       % hyperlinks
\usepackage{url}            % simple URL typesetting
\usepackage{booktabs}       % professional-quality tables
\usepackage{amsfonts}       % blackboard math symbols
\usepackage{nicefrac}       % compact symbols for 1/2, etc.
\usepackage{microtype}      % microtypography
\usepackage{lipsum}		% Can be removed after putting your text content
\usepackage{graphicx}
\usepackage{natbib}
\usepackage{doi}
\usepackage[table]{xcolor}
\usepackage{makecell,multirow}

\usepackage{devanagari}

\usepackage{CJKutf8}

\usepackage{caption}
\usepackage{graphicx}
\graphicspath{ {./figures/} }

\newcolumntype{C}[1]{>{\centering\let\newline\\\arraybackslash\hspace{0pt}}m{#1}}

\title{Lexical Diversity in Kinship Across Languages and Dialects}

\author{
    Hadi Khalilia \\
    DISI \\
    University of Trento \\
    Trento, Italy \\
    \texttt{hadi.khalilia@unitn.it} \\
    \And
    Gábor Bella \\
    Lab-STICC CNRS UMR 628 \\
    IMT Atlantique \\
    Brest, France \\
    \texttt{gabor.bella@imt-atlantique.fr} \\
    \And
    Abed Alhakim Freihat \\
    DISI \\
    University of Trento \\
    Trento, Italy \\
    \texttt{abdel.fraihat@gmail.com} \\
    \And
    Shandy Darma \\
    DISI \\
    University of Trento \\
    Trento, Italy \\
    \texttt{shandy.darma@studenti.unitn.it} \\
    \And
    Fausto Giunchiglia \\
    DISI \\
    University of Trento \\
    Trento, Italy \\
    \texttt{fausto.giunchiglia@unitn.it} \\
}

\date{}

\renewcommand{\headeright}{}
\renewcommand{\undertitle}{}
\renewcommand{\shorttitle}{Lexical Diversity in Kinship Across Languages and Dialects}

\hypersetup{
pdftitle={Lexical Diversity in Kinship Across Languages and Dialects},
pdfauthor={Hadi Khalilia, Gábor Bella, Abed Alhakim Freihat, Shandy Darma, Fausto Giunchiglia},
pdfkeywords={multilingual lexicon, dialect, language diversity, lexical gap, kinship, lexical typology},
}

\begin{document}
\maketitle

\begin{abstract}
Languages are known to describe the world in diverse ways. Across lexicons, diversity is pervasive, appearing through phenomena such as lexical gaps and untranslatability. However, in computational resources, such as multilingual lexical databases, diversity is hardly ever represented. In this paper, we introduce a method to enrich computational lexicons with content relating to linguistic diversity. The method is verified through two large-scale case studies on kinship terminology, a domain known to be diverse across languages and cultures: one case study deals with seven Arabic dialects, while the other one with three Indonesian languages. Our results, made available as browseable and downloadable computational resources, extend prior linguistics research on kinship terminology, and provide insight into the extent of diversity even within linguistically and culturally close communities.
\end{abstract}

% keywords can be removed

\section{Introduction}

The culture and the social structure of a community are reflected in the language spoken by its members. One of the most salient examples of this phenomenon is the worldwide diversity of terms used to describe family structures and relationships. While, thanks to studies such as \citet{murdock1970}, kin terms around the globe are generally well documented, many local variations---across dialects of a single language or across languages of a single country---have not yet been fully described or understood. For example, the term \AR{مَعزوزي} \textit{maazoozi} in the Algerian Arabic dialect, meaning \textit{younger brother}, does not have any equivalent term in the Gulf Arabic dialect. In contrast, the Gulf word \AR{ابن العُود} \textit{ibn alood} meaning \textit{elder brother} does not exist in Algerian, which instead uses the word \AR{سِيدي} \textit{siedi}.

Beyond a linguistic or anthropologic interest, the availability of digital resources on language diversity is also desirable from a computational perspective. Language processing applications need to be aware of such phenomena of diversity in order to provide high-quality results. For example, a machine translation system needs to tackle cases of lexical untranslatability where a word or expression in a source language has no equivalent in a given target language, and the choice of an approximate translation can change the meaning of an utterance. For example, for the English sentence \textit{his cousin gave birth to a twin,} Google Translate provides the Arabic translation \AR{أنجب ابن عمه توأما} \textit{a'njaba ibna a'mihi tawaman} that means \textit{His father's brother's son gave birth to a twin}. This syntactically correct yet unintended meaning of a male giving birth output is due to a \emph{lexical gap}, i.e.~a nonexistent equivalent Arabic term for \emph{cousin}. Such cases of \emph{techno-linguistic bias}---where language technology provides better results \emph{by design} in certain languages than in others---tend to remain hidden in monolingual resources but are revealed in multilingual settings \citep{bella2023towards,bella2022b}.

In recent years, there has been an increasing number of linguistic databases covering a large number of languages. These resources are usually aimed at quantitative studies for comparative linguistics, such as the classification of pain predicates \citep{reznikova2012}, a semantic map of motion verbs \citep{walchli2012}, the modeling of color terminology \citep{mccarthy2019}, the CLICS database of cross-linguistic colexifications \citep{rzymski2020}, DiACL (Diachronic Atlas of Comparative Linguistics), a database for ancient Indo-European languages spoken in Eurasia typology \citep{carling2018}, or the Cross-Linguistic Database of Phonetic Transcription Systems \citep{anderson2018}. Often, such databases use phonetic representations of lexical units or are limited to a few hundred or a few thousand core concepts, limiting their usability for the processing of contemporary written language. In our experience, most of the existing typology-informed NLP research is restricted to exploring language-specific morphosyntactic features and has ignored diversity within lexical resources \citep{batsuren2022}. A notable exception is the Universal Knowledge Core, a massively multilingual lexical database that explicitly represents linguistic diversity and that we reuse in our work.

Our research is part of the \emph{LiveLanguage} initiative, the overarching objective of which is to create, publish, and manage language resources that are ``diversity-aware''---i.e.~that reflect the viewpoints of multiple speaker communities---and that can be reused by multiple communities: linguists, cognitive scientists, AI engineers, language teachers and students \citep{bella2023towards}. Contrary to mainstream exploitative practices, LiveLanguage aims to carry out its goals while empowering local speaker communities, giving them control over resources they help to produce \citep{helm2023diversity}. Involving human contributors and deciders from speaker communities is therefore a crucial part of our methodology. 

In particular, the present paper focuses on diversity where it is less expected to appear: within dialects of the same language and within languages of the same country. Therefore, we describe a multidisciplinary study on the diversity of kin terms across seven Arabic dialects (Algerian, Egyptian, Tunisian, Gulf, Moroccan, Palestinian, and Syrian) and three languages from Indonesia (Indonesian, Javanese, and Banjarese). We consider kin terms as a domain particularly well-suited both for research on the methodology of collecting and producing diversity-aware linguistic data, and for comparative studies on diversity across languages. 

Our paper aims to provide four contributions: (1)~a general method for collecting multilingual lexical data from native speakers for a given domain (in our case the domain of kin terms), in a diversity-aware manner; (2)~223~kin terms and 1,619~lexical gaps collected in seven Arabic dialects and three Indonesian languages; (3)~a qualitative and quantitative discussion of our results regarding the diversity observed across the dialects and languages covered; and (4)~the publication of our results as an open, computer-processable dataset, as well as its integration into the Universal Knowledge Core multilingual database. Our starting point is state-of-the-art datasets on worldwide kinship terminology from ethnography \citep{murdock1970} and computational linguistics \citep{Khishigsuren2022}. Our data collection method is based on collaborative input from native speakers and language experts. Our results extend the state-of-the-art resources above with kin terms in languages and dialects not yet covered, as well as with 22~new kinship concepts not yet associated with other languages within those resources.

The structure of the paper is organized as follows. In Section \ref{secTypology} we give an overview of lexical typology and the phenomena of lexical untranslatability and lexical gaps with respect to the domain of kinship in particular. The Universal Knowledge Core resource is presented in Section \ref{secUKC}. In Section \ref{secMethod} we describe our data collection method. Sections \ref{secArabic} and \ref{secIndonesian} introduce two case studies on Arabic dialects and Indonesian languages, respectively. Section \ref{secRelatedWork} discusses previous studies related to our work. Finally, we provide conclusions in Section \ref{secConclusion}.

\section{Untranslatability and Lexical Typology}
\label{secTypology}

Linguists understand translation from one language to another as a complex and multidimensional problem, ranging from multiple coexisting forms of meaning equivalence to untranslatability \citep{catford1965, bella2022b}. The diversity between cultures is a major cause for this problem appearing on several lexical-semantic levels. Some examples of the linguistic diversity are the richness of Toaripi vocabulary on the various forms of motion verbs describing walking around the beach like (isai) meaning ``\textit{go beachward}'' and (kavai) meaning ``\textit{go inland with respect to the beach}'', the language of the coastal Papua New Guinea country, the lack of vocabulary for the word meaning ``\textit{sailing}'' in Mongolian, which is the language of a landlocked country, or the Arabic word \AR{تَسنَّم} meaning ``\textit{to ascend a camel's hump}''.

The domain of kinship terms, which is the subject of our paper, is known to be extremely varied across languages, due to the different ways family structures are organised around the world. Matriarchal societies may describe certain female relatives with more detail, while strongly patriarchal ones are more descriptive with respect to male relatives. Arabic dialects, for instance, distinguish paternal and maternal brothers but also blood brothers, full brothers, and breastfeeding brothers. Thus, not only are kinship-related vocabularies `richer' or `poorer' across languages, they are also structured in different manners.

In this research, we focus on lexical untranslatability, which manifests most clearly through the lexical gap phenomenon when a word in a source language does not have a concise and precise translation in a given target language. Lexical gaps are often the linguistic manifestation of culturally or spatially defined specificities of a community of language speakers that cannot entirely be predicted or explained through systematic principles or recurrent patterns \citep{lehrer1970}. Table \ref{Table1} below presents this phenomenon for nine concepts representing sibling relationships from the kinship domain in eight languages\footnote{These nine concepts do not cover sibling terms exhaustively in all languages: for example, many Austronesian languages use different terms based on the gender of the speaker.}.
One can observe that none of the eight languages has concise lexicalizations for all nine concepts, yet each concept is lexicalized in at least one language. Such variations in lexicalization pose a problem for both machine and human translation: for instance, substituting a specific term instead of a broader one may result in injecting unintended meaning. In Javanese, at least four specific terms---(sedulur/\textit{sibling}), (adhi/\textit{younger sibling}), (kangmas/\textit{elder brother}), and (Mbakyu/\textit{elder sister})---are used for expressing the sibling relationship, and accordingly, translating this sentence through Google Translate (\textit{my sister is ten years older than me}) to Javanese gives this nonsensical sentence (\textit{adhiku luwih tuwa sepuluh taun tinimbang aku}) meaning (\textit{my younger sibling is ten years older than me}). This result is due to the lack of Javanese vocabulary for the word meaning (sister), and also lacks the term meaning ``\textit{younger sister}'', so the machine translator uses (adhi) meaning ``\textit{younger sibling},'' which finally produces the semantically absurd output.

\begin{table}[h!]
    \footnotesize
    \renewcommand{\arraystretch}{1.7}
    \caption{Lexicalizations of nine meanings around the concept of (sibling) in eight languages. \\}
    \centering
    \begin{tabular}{ |p{0.8in}|c|c|c|C{0.5in}|C{0.6in}|c|c|c| }
        \hline
        \textbf{Meaning} & \textbf{English} & \textbf{Japanese} & \textbf{Arabic} & \textbf{Italian} & \textbf{Indonesian} & \textbf{Hindi} & \textbf{Hungarian} & \textbf{Javanese} \\
        \hline
        sibling & sibling & \cellcolor[HTML]{D3D3D3} GAP & \cellcolor[HTML]{D3D3D3} GAP & \cellcolor[HTML]{D3D3D3} GAP & saudara & {\dn shodr} & testvér & sedulur \\
        \hline
        elder sibling & \cellcolor[HTML]{D3D3D3} GAP & \cellcolor[HTML]{D3D3D3} GAP & \cellcolor[HTML]{D3D3D3} GAP & \cellcolor[HTML]{D3D3D3} GAP & kakak & \cellcolor[HTML]{D3D3D3} GAP & nagytestvér & \cellcolor[HTML]{D3D3D3} GAP \\
        \hline
        younger sibling & \cellcolor[HTML]{D3D3D3} GAP & \cellcolor[HTML]{D3D3D3} GAP & \cellcolor[HTML]{D3D3D3} GAP & \cellcolor[HTML]{D3D3D3} GAP & adik & \cellcolor[HTML]{D3D3D3} GAP & kistestvér & adhi \\
        \hline
        brother & brother & \cellcolor[HTML]{D3D3D3} GAP & \AR{أخ} & fratello & \cellcolor[HTML]{D3D3D3} GAP & \dn{B\4yA} & \cellcolor[HTML]{D3D3D3} GAP & \cellcolor[HTML]{D3D3D3} GAP \\
        \hline
        sister & sister & \cellcolor[HTML]{D3D3D3} GAP & \AR{أُخْت} & sorella & \cellcolor[HTML]{D3D3D3} GAP & \dn{bhn} & \cellcolor[HTML]{D3D3D3} GAP & \cellcolor[HTML]{D3D3D3} GAP \\
        \hline
        elder brother & \cellcolor[HTML]{D3D3D3} GAP & \begin{CJK}{UTF8}{min}あに\end{CJK} & \cellcolor[HTML]{D3D3D3} GAP & fratellone & abang & \dn{B\4yA} & báty & kangmas \\
        \hline
        elder sister & \cellcolor[HTML]{D3D3D3} GAP & \begin{CJK}{UTF8}{min}あね\end{CJK} & \cellcolor[HTML]{D3D3D3} GAP & sorellona & \cellcolor[HTML]{D3D3D3} GAP & \dn{dFdF} & nővér & mbakyu \\
        \hline
        younger brother & \cellcolor[HTML]{D3D3D3} GAP & \begin{CJK}{UTF8}{min}おとうと\end{CJK} & \cellcolor[HTML]{D3D3D3} GAP & fratellino & \cellcolor[HTML]{D3D3D3} GAP & \dn{BAI} & öcs & \cellcolor[HTML]{D3D3D3} GAP \\
        \hline
        younger sister & \cellcolor[HTML]{D3D3D3} GAP & \begin{CJK}{UTF8}{min}いもうと\end{CJK} & \cellcolor[HTML]{D3D3D3} GAP & sorellina & \cellcolor[HTML]{D3D3D3} GAP & \dn{bhn} & húg & \cellcolor[HTML]{D3D3D3} GAP \\
        \hline
    \end{tabular}
    \label{Table1}
\end{table}

Lexical typology is a field of linguistics that studies the diversity across languages according to the structural features of languages with respect to specific semantic fields \citep{plungyan2011}. Different classical studies are conducted in this field on grammar and phonology, such as VoxClamantis V1.0― a large-scale corpus for phonetic typology \citep{salesky2020} and the structure of the space semantic field by identifying a set of semantic parameters and notions depending on the grammatical information of the field's constituents \citep{levinson2006}. Other examples of such studies have been conducted on lexical-typological issues that appear across languages during translation, like the presence or absence of lexicalizations in languages. In these articles, authors focused on semantic fields that offer the richness of cross-lingual diversity: family relationships \citep{kemp2012}, colors \citep{roberson2005}, food \citep{bella2022a}, body parts \citep{wierzbicka2007}, putting and taking events \citep{kopecka2012}, cutting and breaking events \citep{majid2007}, or cardinal direction terms \citep{arora2021}. 
However, as mentioned in the introduction, only a few open datasets have been published in the scientific research area. These include the classification of kinship by \citet{murdock1970}, which has been published in D-PLACE \citep{kirby2016}. Part of \citet{kay2016}'s work on colors is published under the lexicon chapter of the World Atlas of Language Structures (WALS) \citep{dryer2013}. Additionally, a color categorization dataset by \citet{mccarthy2019} is available on GitHub\footnote{\url{https://github.com/aryamccarthy/basic-color-terms}}.

Digital lexicons have been increasingly used in lexical typology, enabling typologists to explore a broader range of languages and semantic domains. One noteworthy example is the KinDiv\footnote{\url{http://ukc.disi.unitn.it/index.php/kinship}} lexicon \citep{Khishigsuren2022}, which encompasses 1,911 words and identifies 37,370 gaps within the domain of kinship, spanning 699 languages. In our current research, we extend our investigation into the kinship domain, specifically focusing on exploring linguistic diversity among Arabic dialects and Indonesian languages. Other examples include \citet{viberg1983}'s seminal study, which was conducted on perceptual terminology in 50 languages and has been expanded upon by \citet{georgakopoulos2022} to cover 1,220 languages. Furthermore, the Kinbank database, recently introduced by \citet{kinbank2023}, serves as a comprehensive repository of kinship terminology, encompassing more than 1,173 languages and offering a broad coverage of various kinship subdomains.

\section{Universal Knowledge Core}
\label{secUKC}

\begin{figure}
    \centering
    \includegraphics[scale=0.19]{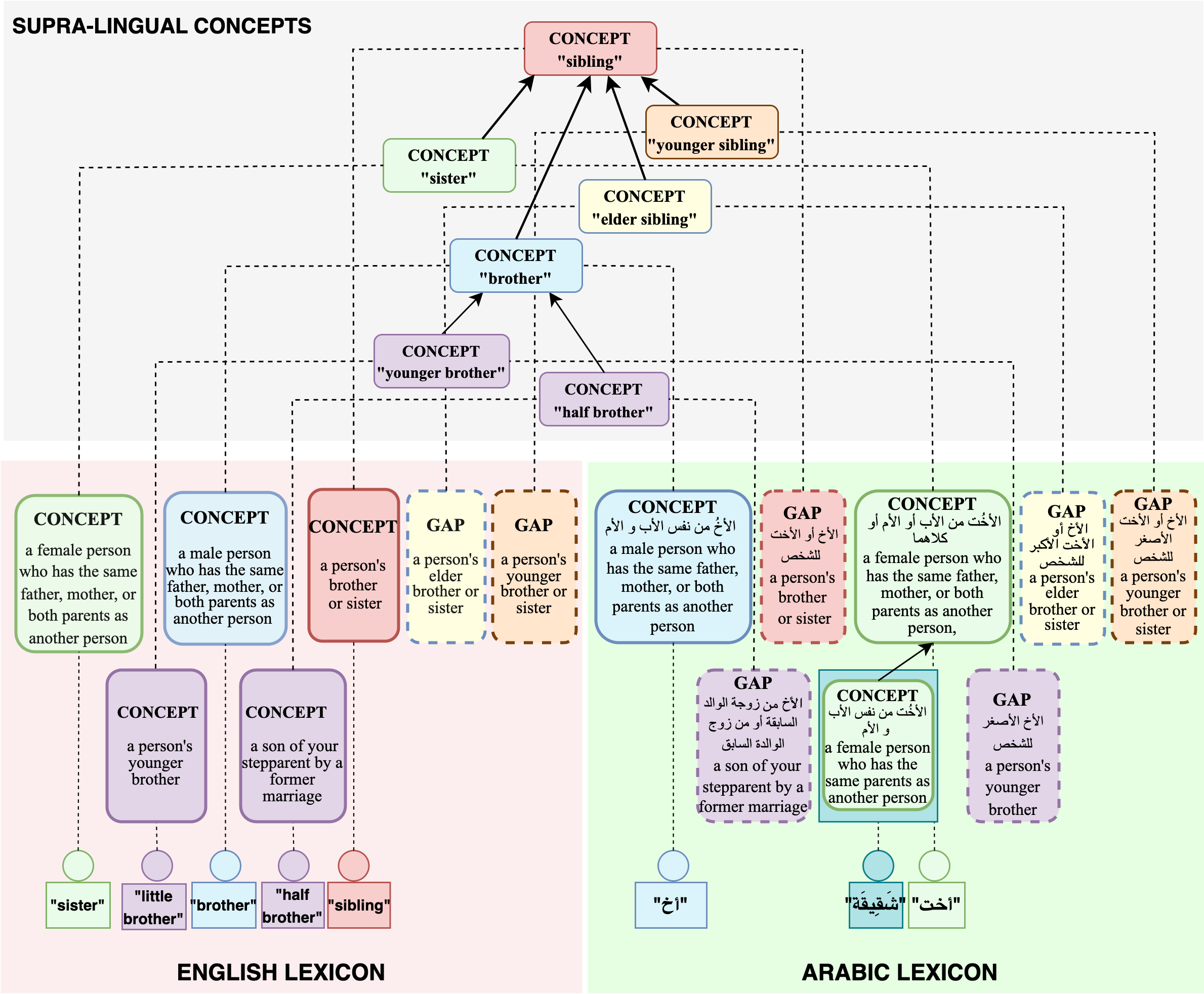}
    \caption{Structural elements in the UKC lexical database}
    \label{fig1}
\end{figure}

This section describes the Universal Knowledge Core (UKC)\footnote{\url{http://ukc.datascientia.eu}}, a large multilingual lexical database that we adopt for the production of diversity-aware datasets in this research \citep{giunchiglia2017}. The use of the UKC is motivated by its ability to represent linguistic unity and diversity explicitly: conceptualisations shared across languages, word senses appearing only in certain languages, shared lexicalisations (e.g.~cognates), as well as lexical gaps.
The theoretical underpinnings of the lexical model of the UKC have been described in \citet{giunchiglia2018} and in \citet{bella2022a}, and are illustrated in Figure \ref{fig1}.

The UKC is divided into a supra-lingual concept layer (as shown at the top of Figure \ref{fig1}) and the layer of individual lexicons (at the bottom of Figure \ref{fig1}). The concept layer includes hierarchies of concepts that represent lexical meaning shared across languages. Concepts are language-independent units and act as bridges across languages, and each one should be lexicalized by at least one language to be present in the concept layer. Supra-lingual concepts and their relations (e.g.~hypernymy, meronymy) are in part derived from third-party resources such as Princeton WordNet (PWN) \citep{miller1995}, and are in part proper to the UKC. In particular, the UKC contains an extensive formal conceptualisation of kinship domain terms computed from the KinDiv database, spanning about 200~distinct concepts.\footnote{\url{https://github.com/kbatsuren/KinDiv}} KinDiv itself is based on ethnographic evidence from 699~languages \citep{Khishigsuren2022}. While this existing hierarchy of kinship concepts does not fully cover all terms that appear in our study, it is the most complete one we are aware of, motivating our choice of the UKC as a platform for our research.

The lexicon layer consists of language-specific lexicons that provide lexicalizations for the concepts from the supra-lingual concept layer, while also asserting \emph{lexical gaps} whenever lexicalizations are known not to exist. Lexicons also provide term definitions as well as lexical relationships specific to the language, such as derivations, metonymy, or antonymy relations. Lexicons can also contain \emph{language-specific concepts} that do not appear in the supra-lingual concept layer. For example, in Figure \ref{fig1}, the Arabic \AR{شَقِيقة}, meaning ``\textit{a female person who has the same father, mother, or both parents as another person}'', is represented as a language-specific concept. The dual mechanism of defining lexical concepts either on the supra-lingual or on the language-specific level allows for the representation of differing worldviews that would be hard or impossible to reconcile into a single global concept graph. The richness of its lexicon-level linguistic knowledge makes the UKC unique among multilingual lexical databases and particularly suitable for our study.

As mentioned in Section~\ref{secTypology}, a lexical gap for a specific concept is present in a language if there is no concise equivalent word meaning for the concept in that language. For example, neither English nor Arabic has a word meaning \emph{elder sibling}; for such cases, the UKC provides evidence of meaning non-existence and untranslatability by representing lexical gaps inside lexicons, as shown in Figure \ref{fig1}. This information can be used by the NLP community to indicate the absence of equivalent words to downstream cross-lingual applications.

Beyond providing lexical relations between shared word meanings as other multilingual lexical databases do, the UKC also represents a richer set of lexical-semantic connections between language units in a lexicon. For example, the \textit{antonym} lexical relation expresses that two senses are opposite in meaning. While the lexical-semantic relation, \textit{similar-to}, is used to connect two concepts with similar meanings, and the \textit{hypernym-of} connects parent meaning with its child. For instance, in Figure \ref{fig1}, the English (little brother) and (brother) are connected through a \textit{hypernym-of} relationship. Such information can be used by the NLP community to indicate the concise equivalent language-specific word meaning to downstream cross-lingual applications, e.g., as the position of a language-specific meaning in a language hierarchy in a lexicon.

The UKC currently does not explicitly distinguish between languages and dialects: each vocabulary is a separate entity labeled with a standard three-letter ISO~639-3 code. When such a code is not available, the UKC uses a standard extension mechanism where three additional (not standardised) letters are added to the ISO code: e.g.,~for Syrian Arabic, the code \texttt{arb-syr} is used.

\section{A Methodology for Building Diversity-Aware Lexicons}
\label{secMethod}

This section presents the general method by which we collected and produced lexicalizations and gaps from native speakers and language experts. The same method presented below was employed in an independent manner for each Arabic dialect and Indonesian language covered by our study. The contents of this section aim to serve as a tried and tested recipe for gathering lexical data in a diversity-aware manner, that we intend to reuse in future lexicon development projects.

We exploit the UKC to import language-independent concepts (e.g., kinship concepts) to be used as an input dataset to our method and use its data representation model to formalize our data. We reuse an already broad and well-formalised hierarchy of 184~kinship concepts from the KinDiv database, which includes kinship terms and gaps in 699~languages. Data in KinDiv is based on the well-known results of \citet{murdock1970}, as well as on lexicalizations retrieved from Wiktionary that we consider as an overall good-quality resource. In \citet{Khishigsuren2022}, the accuracy of KinDiv was evaluated to be above 96\%. One language expert per language provided this percentage, which represents the proportion of the number of words (or gaps) validated as correct to the total number of collected words (or gaps).

\begin{figure}[h]
    \centering
    \includegraphics[scale=0.53]{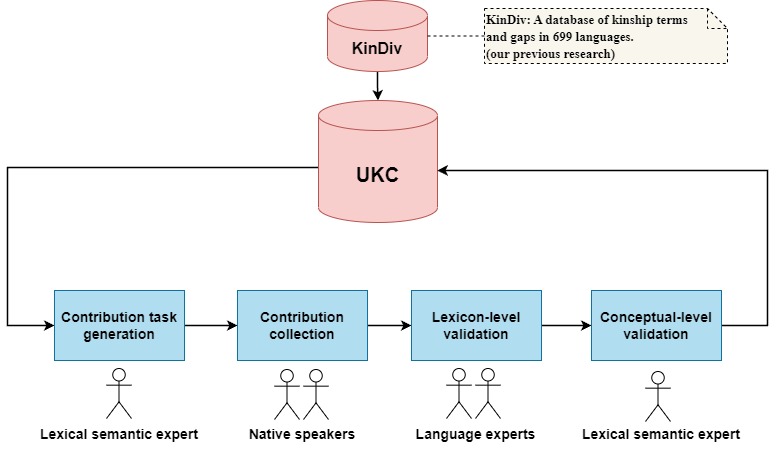}
    \caption{Methodology macro-steps and data sources}
    \label{fig2}
\end{figure}

Our work extends KinDiv data by new concepts, lexicalizations, and lexical gaps in languages and dialects that are either not present in KinDiv or are incompletely covered. A lexical-semantic expert generates a contribution (kinship terms and gaps) task, then a group of native speakers collects contributions from a dialect (and a local language). After that, two steps for validating collected contributions: language experts evaluate collected lexical units and gaps of a dialect, and a lexical-semantic expert evaluates explored kinship concepts (not existing in UKC). Additionally, resulting data (including gaps, words, and new concepts) is used to update and enrich UKC. So, gaps and words are merged into the lexicons of the UKC  while new concepts are integrated with the (top) concept layer. A general view of the method is depicted in Figure \ref{fig2}. 

Accordingly, the macro-steps of our methodology are as follows:

\begin{enumerate}
    \item \textit{Contribution task generation}: First, prepare the materials: the dataset of inputs to be examined and the architecture of the supra-lingual concept layer of each subdomain.
    \item \textit{Contribution collection}: The actual contribution effort is carried out by a native speaker in a local language or dialect.
    \item \textit{Lexicon-level validation}: Provided words and gaps are evaluated and corrected by a language expert.
    \item \textit{Concept-level validation}: New concepts and unclear contributions (i.e., words on the borderline) are verified by a lexico-semantic expert.
\end{enumerate}

\subsection{Contribution task generation}
This section describes the material needed during the execution of the next steps of the methodology. Hence, two constituents must be prepared in this step as described below:

\begin{enumerate}
    \item \textit{Dataset of inputs}: Constructing the dataset of general word meanings is the first step of studying diversity across dialects and represents the inputs of the contribution collection phase. In this context, the UKC lexicon is employed to build a dataset, which contains several facilities that support retrieving categorized data from its interlingual shared meaning layer as introduced in Section \ref{secUKC}. Moreover, typology datasets or other approaches can be used for that, such as the kinship dataset from \citet{murdock1970}; or gathering data from online dictionaries using automatic methods, i.e.~KinDiv retrieves some of its kinship terms from Wiktionary. The constructed dataset is a spreadsheet containing language-independent meanings from one semantic field. At the same time, its content is distributed into subdomains (sheets) for usability and simplicity in designing a concept hierarchy for each subdomain which is a helpful tool for lexical-gap exploration. One spreadsheet row is generated for each concept, containing the concept ID, the source concept definition in the standard language, another definition in English, as well as empty slots for inserting a lexical gap or a word with equivalent meaning, and the data provider's comments in a dialect or local language.
    \item \textit{Interlingual concept hierarchy}: Modeling the interlingual shared meaning space is essential to explore lexical gaps systematically. In this task, the UKC concept hierarchy is exploited. UKC is the only resource introducing a hierarchy of shared meanings across languages for each semantic field, such as kinship, colors, or food. Furthermore, UKC uses a hybrid linguistic-conceptual approach in modeling each domain. This approach adopts actual domain ontology and linguistic data from typological literature. For example, a fragment of the brotherhood hierarchy in the top layer of the UKC is shown in Figure \ref{fig1}. A native speaker can compare each examined concept from the spreadsheet with the hierarchy of its domain to extract additional knowledge about its meaning based on a concept's position in the hierarchy, which helps to provide a concrete answer in terms of a gap or a lexical unit.
\end{enumerate}

\subsection{Contribution collection}

Contributions from a local language or a dialect are provided by one native speaker who was born and educated (university level) within the speaker community. The following are the most notable instructions they are given:

\begin{enumerate}
    \item They are given the authority to skip concepts, stop contributions, or leave a comment when they deem the terms are becoming too culture-specific and consequently need an exact answer.
    \item They are asked to provide a lexicalization in a local language (or dialect) that gives meaning equal to the concept's meaning.
    \item They are asked explicitly to identify lexical gaps where no local (or dialect) lexicalization exists.
    \item Within a local language (or dialect) and a subdomain (e.g., cousins), they are asked to provide new concepts that did not exist in the list of inputs which is imported from the UKC by providing a word (lemma) and a clear description of its meaning.
\end{enumerate}

\begin{figure}[h]
    \centering
    \includegraphics[scale=0.1]{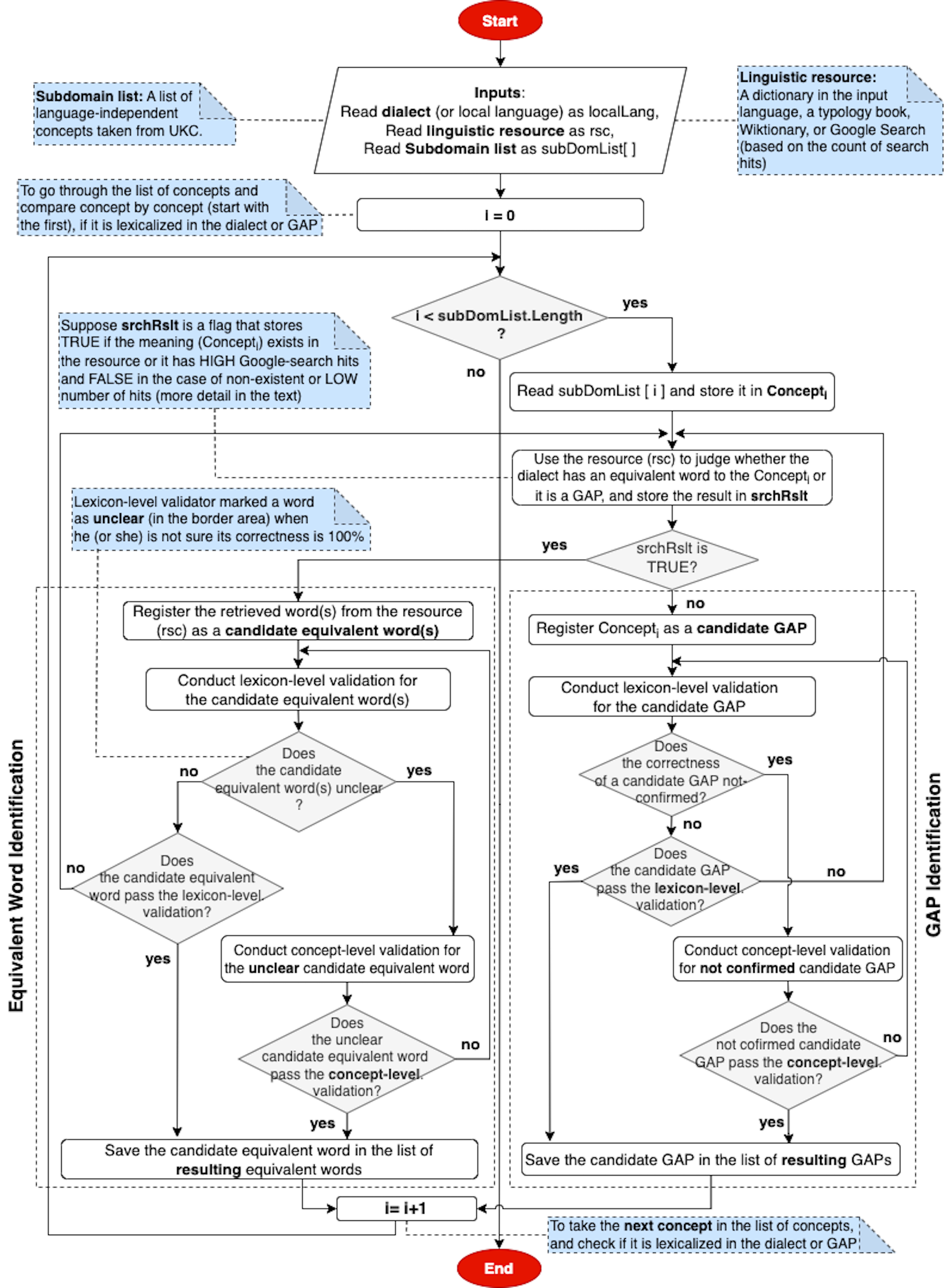}
    \caption{Flowchart of gap and equivalent word meaning identification}
    \label{fig3}
\end{figure}

The process of providing such contributions is depicted in two flowcharts; for instance, Figure \ref{fig3} shows the flowchart of the candidate gap (on the left-hand side of the figure) and candidate equivalent word meaning (on the right-hand side of the figure) exploration; it starts identifying a standard language and a local language (or dialect) and providing a native speaker with a spreadsheet including a list of subdomain concepts (inputs). Then, a native speaker is asked to find a linguistic resource in the local language and use it to search for concepts (concept-by-concept) to confirm lexicalizations and gaps. He/she can use a linguistic resource in the search process as the following steps: searching in a well-known dictionary, then in Wiktionary— a large multilingual online lexicon after that in a typology dataset (if it is available), and finally, using Google search (based on the count of search hits). More details about these steps are described in Section \ref{secArabic}. The native speaker can rely on search results and the count of Google hits to give a more concrete answer on whether the concept in the standard language has a lexicalization or is a gap in the local language; such candidates are passed to the next phase- lexicon-level validation.

\begin{figure}[h]
    \centering
    \includegraphics[scale=0.1]{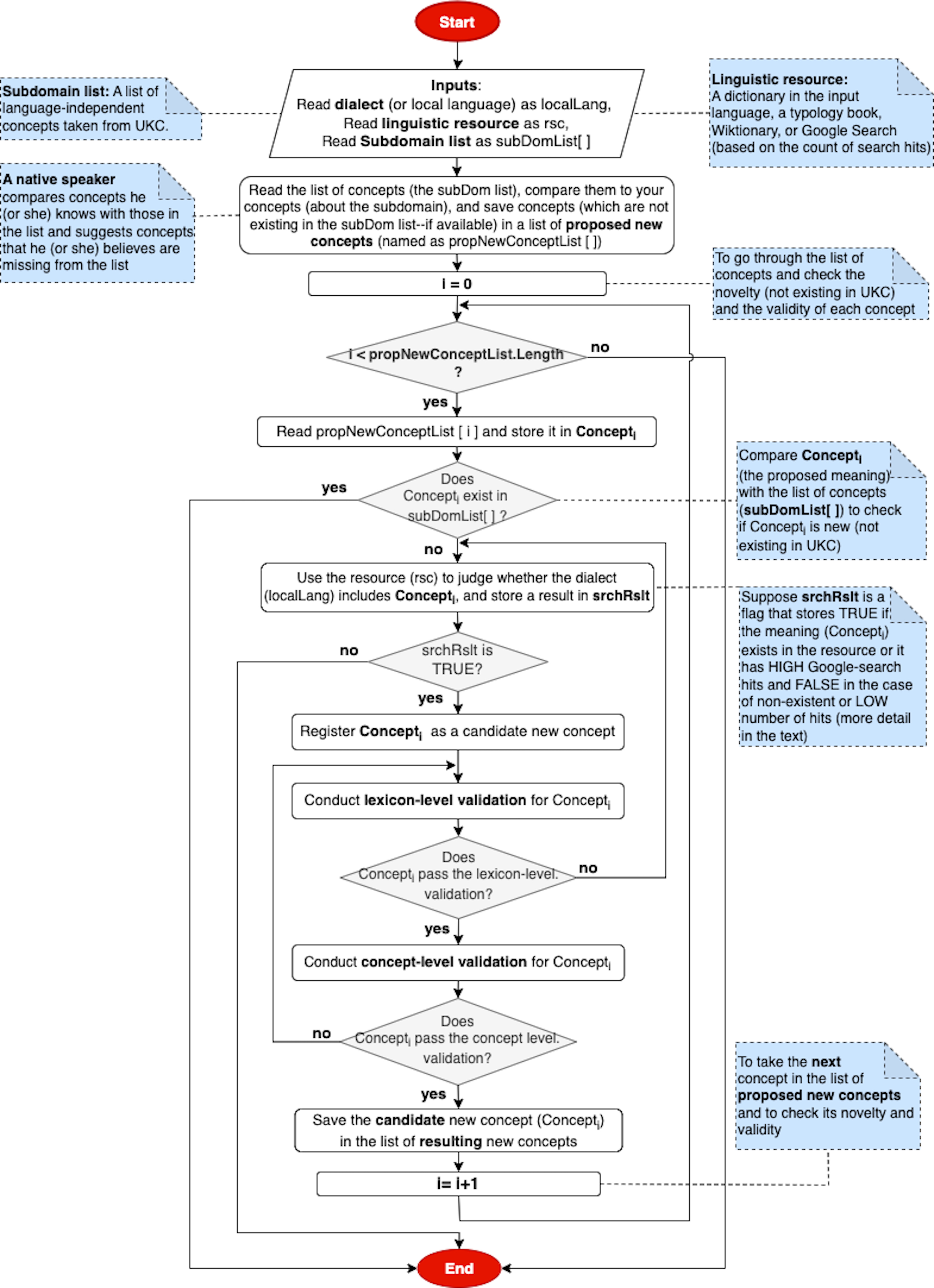}
    \caption{Flowchart of a new concept collection}
    \label{fig4}
\end{figure}

A new concept collection is a third contribution in this phase, where the steps of a candidate new concept exploration in a local language can be seen in Figure \ref{fig4}. A native speaker can examine the list of subdomain concepts and provide his/her (own) concepts with their definitions that he/she believes have not existed in the list. The same search steps in gap identification can be followed in this task. As shown in Figure \ref{fig4}, All candidate new concepts are passed to the two subsequent validation phases: lexicon- and conceptual level.

\subsection{Lexicon-Level Validation}

Our lexicon-level validation method formally and explicitly addresses individual gap identifications and their quality, as well as equivalent word meanings and new concepts. It allows a qualitative evaluation of the entire list of provided contributions through word-by-word and gap-by-gap in a loop between a native speaker and a validator. A word, a gap, or a new concept does not pass this validation until the native speaker provides the correct answer for each of them, as shown in the flowcharts in Figure \ref{fig3} and Figure \ref{fig4}.

A language expert who is also a native speaker of the determined language (or dialect) will carry out this validation on a spreadsheet containing the data and results gathered in the previous step with two additional empty columns: the evaluation and lexicon-level validator's comment, producing the following information:

\begin{enumerate}
    \item \textit{Equivalent word meanings}: validate the correctness of all provided words in the local language (or dialect) by marking them up as correct, incorrect, or unclear for borderline cases and by providing correct words or indicating them as lexical gaps for incorrect ones.
    \item \textit{Lexical gaps}: validate the word meanings marked as lexical gaps by the native speaker in the local language, either as confirmed gaps or as non-gaps due to an existing lexicalization in that language, which the validator needs to indicate.
    \item \textit{New concepts}: validate all proposed new word meanings in each subdomain by marking them up as correct, correct but not new (in case the supposedly new concepts already existed in the list), or not accepted (in case another concept already existed in the list to express the meaning, or the validator does not consider it as a desirable suggestion for other reasons).
\end{enumerate}

Correct equivalent word meanings and gaps are integrated with the local language lexicon on the fly. Also, correct new concepts are passed to the next step to be validated at the concept level before merging them with the supra-lingual shared meaning layer. While in case the evaluation is an incorrect equivalent word or a gap, or not accepted new concept, the validator returns each of them with a comment describing the reason to the native speaker to review and address the problem; when the native speaker finishes revising them, then he/she returns the new version of a contribution to the validator. This cycle (native speaker's contribution---lexicon level validation) is still alive until the validator confirms the correctness of the contribution or skips it.

\subsection{Concept-Level Validation}

In this step, a lexical-semantic expert who is the manager of the UKC system verifies the new concepts and their quality as accept or reject to add them into the supra-lingual concept layer as well as addresses unclear words and non-confirmed gaps/non-gaps that are borderline cases. This validation is based on a discussion session with the language expert responsible for lexicon-level validation through concept-by-concept and case-by-case issue validation. A spreadsheet containing all new concepts and determined (words and gaps) to be examined is used. Columns of this sheet are the same columns in the previous step and two additional empty ones: the evaluation and concept-level validator's comment. The following tasks are used:

\begin{enumerate}
    \item \textit{New concepts}: Validate all proposed new concepts in each subdomain by marking them up as correct, correct but not new (in case the supposedly new concepts already existed in the UKC), or not accepted (in case another concept already existed in the UKC to express the meaning, or the validator does not consider the new concept as a desirable suggestion for any other reason).
    \item \textit{Unclear words}: Validate the correctness of unclear word cases considered in the border-area by the lexicon-level validator by marking them as correct or incorrect and writing a comment.
    \item \textit{Non-confirmed gaps/non-gaps}: Validate the word meanings that do not have confirmation as lexical gaps or non-gaps by providing a judgment with a comment
\end{enumerate}

Correct new concepts are imported into UKC by merging them with the supra-lingual conceptual layer. In contrast, not-accepted ones and those correct but not new are returned to the validator at the lexicon level, who may also return them with a comment describing the reason to the native speaker to address an included problem. In a new cycle, modified new concepts by the native speaker are transferred to this phase through the validator of lexicon-level; then, the validator at this level reviews the updates and decides whether to finish the revision cycle by accepting or rejecting the new concepts or issue a new one for more review, as shown in Figure \ref{fig4}. In addition, confirmed words and gaps output from this step are integrated with the language lexicon in the UKC, as shown in Figure \ref{fig3}. 

\section{Case Study on Diversity Across Arabic Dialects}
\label{secArabic}

This section demonstrates the use of the methodology described in Section~\ref{secMethod} on kinship terminology from seven dialects of the Arabic language.
Arabic is the official language of more than four hundred million native speakers in twenty-two countries in the Middle East and northern Africa. Classical Arabic or Modern Standard Arabic (MSA) refers to the standard form of the language used in academic writing, formal communication, classical poetry, and religious sermons \citep{elkateb2006}. Surprisingly lexical diversity is manifested between Arabic dialects, evident in our study between seven of the twenty dialects spoken worldwide. The selected dialects are Egyptian, Moroccan, Tunisian, Algerian, Gulf, and South Levantine (two examples: Palestinian and Syrian). Let us take the example of the Gulf word \AR{الخَال العودْ} meaning ``\textit{mother's elder brother},'' which has no equivalent in South Levantine or Moroccan; instead, they use the more general word \AR{الخَال} meaning ``\textit{mother's brother},'' which can be used for both meanings ``\textit{mother's younger brother}'' or ``\textit{mother's elder brother}''. In this paper, we perform an experiment on the Arabic dialects to capture their diversity in the kinship domain. The resulting dataset with dialect-specific kinship terms will be integrated with an instance of the Universal Knowledge Core for Arabic (Arabic UKC)\footnote{\url{http://arabic.ukc.datascientia.eu/concept}} ongoing project, which is the first diversity-aware lexical resource for Arabic dialects so far.

\subsection{Experiment Setup}

As mentioned in Section \ref{secUKC}, the UKC resource is our data source in building the input dataset of kinship-independent language concepts and formalizing such concepts and new word meanings (not existing in the inputs) explored in this experiment. For example, the brotherhood hierarchy is shown in the top layer of the UKC in Figure \ref{fig1}. In this study, contributions are provided by seven native speakers (one per Arabic dialect). Regarding the contributors' socio-linguistic background, each has at least a master's degree and was born and educated, at least up to high school level, within the native speaker community. The participants' linguistic backgrounds are presented below:

\begin{enumerate}
    \item \textit{Participant 1}: a native Algerian speaker with good command of English.
    \item \textit{Participant 2}: a native Egyptian speaker with good command of English.
    \item \textit{Participant 3}: a native Tunisian speaker with good command of English and French.
    \item \textit{Participant 4}: a native Gulf speaker with good command of English and Arabic-Palestinian.
    \item \textit{Participant 5}: a native Moroccan speaker with good command of English and Italian.
    \item \textit{Participant 6}: a native Palestinian speaker with good command of Arabic-Syrian and English.
    \item \textit{Participant 7}: a native Syrian speaker with good command of English.
\end{enumerate}

Seven experiments (one for each dialect) are performed to explore lexical units and gaps using our method. In each experiment, a spreadsheet of kinship concepts is imported from the UKC (as the source, they were computed from the KinDiv database), which serves as an input dataset to the contribution (diversity-aspects) collection step. These kinship domain concepts are language-independent units representing lexical meaning shared across 699 languages and spanning 184 distinct concepts. UKC categorizes kinship concepts into six groups; each one contains a distinct subset of concepts sharing a common kinship type meaning called a subdomain, for example, sibling and cousin subdomains. The spreadsheet (the dataset) consists of six sheets, and each one represents a kinship subdomain. See Table \ref{Table2}, which shows the subdomain names and the count of containing concepts per subdomain of the dataset.

\begin{table}[h!]
    \caption{Statistics on the concepts in the input dataset. \\}
    \renewcommand{\arraystretch}{1.5}
    \centering
    \begin{tabular}{ |c|c| }
        \hline
        \textbf{Subdomains} & \textbf{Count of Concepts} \\
        \hline
        Grandparents & 19 \\
        \hline
        Grandchildren & 27 \\
        \hline
        Siblings & 21 \\
        \hline
        Uncle/Aunt & 27 \\
        \hline
        Nephew/Niece & 33 \\
        \hline
        Cousins & 57 \\
        \hline
        \textbf{Total} & 184 \\
        \hline
    \end{tabular}
    \label{Table2}
\end{table}

\begin{table}[]
    \caption{Count of Google Search Hits for Cousin Concepts in Arabic. \\}
    \centering
    \renewcommand{\arraystretch}{1.7}
    \begin{tabular}{|c|c|cc|}
        \hline
        \textbf{Concept} & \textbf{With/Without Diacritics} & \multicolumn{2}{c|}{\textbf{Count of Hits}} \\
        \hline
        \multirow{2}{*}{\makecell{\AR{العمومة} \\ Paternal cousin}} & \AR{العُمومَةُ} & \multicolumn{1}{l|}{1.94 M} & \multirow{2}{*}{3.04 M} \\
        \cline{2-3}
        & \AR{العمومة} & \multicolumn{1}{l|}{1.1 M} & \\
        \hline
        \multirow{2}{*}{\makecell{\AR{الخؤولة} \\ Maternal cousin}} & \AR{الخُؤولَةٌ} & \multicolumn{1}{l|}{111 k} & \multirow{2}{*}{158 k} \\
        \cline{2-3}
        & \AR{الخؤولة} & \multicolumn{1}{l|}{47 k} & \\
        \hline
        \multirow{2}{*}{\makecell{\AR{ابن العم} \\ Son of father's brother}} & \AR{اِبْن العَم} & \multicolumn{1}{l|}{84.8 M} & \multirow{2}{*}{93.96 M} \\
        \cline{2-3}
        & \AR{ابن العم} & \multicolumn{1}{l|}{9.16 M} & \\
        \hline
        \multirow{2}{*}{\makecell{\AR{بِنت العم} \\ Daughter of father's brother}} & \AR{بِنْت العَم} & \multicolumn{1}{l|}{8.43 M} & \multirow{2}{*}{83.13 M} \\
        \cline{2-3}
        & \AR{بِنت العم} & \multicolumn{1}{l|}{74.7 M} & \\
        \hline
        \multirow{2}{*}{\makecell{\AR{ابن العمة} \\ Son of father's sister}} & \AR{اِبْن العَمَّة} & \multicolumn{1}{l|}{12.5 M} & \multirow{2}{*}{131.5 M} \\
        \cline{2-3}
        & \AR{ابن العمة} & \multicolumn{1}{l|}{119 M} & \\
        \hline
        \multirow{2}{*}{\makecell{\AR{بِنت العمة} \\ Daughter of father's sister}} & \AR{بِنْت العَمَّة} & \multicolumn{1}{l|}{9 M} & \multirow{2}{*}{30.4 M} \\
        \cline{2-3}
        & \AR{بِنت العمة} & \multicolumn{1}{l|}{21.4 M} & \\
        \hline
        \multirow{2}{*}{\makecell{\AR{ابن الخال} \\ Son of mother's brother}} & \AR{اِبْن الخَال} & \multicolumn{1}{l|}{5.61 M} & \multirow{2}{*}{33.01 M} \\
        \cline{2-3}
        & \AR{ابن الخال} & \multicolumn{1}{l|}{27.4 M} & \\
        \hline
        \multirow{2}{*}{\makecell{\AR{بِنت الخال} \\ Daughter of mother's brother}} & \AR{بِنْت الخَال} & \multicolumn{1}{l|}{3.99 M} & \multirow{2}{*}{30.69 M} \\
        \cline{2-3}
        & \AR{بِنت الخال} & \multicolumn{1}{l|}{26.7 M} & \\
        \hline
        \multirow{2}{*}{\makecell{\AR{ابن الخالة} \\ Son of mother's sister}} & \AR{اِبْن الخَالَة} & \multicolumn{1}{l|}{12.5 M} & \multirow{2}{*}{16.59 M} \\
        \cline{2-3}
        & \AR{ابن الخالة} & \multicolumn{1}{l|}{4.09 M} & \\
        \hline
        \multirow{2}{*}{\makecell{\AR{بِنت الخالة} \\ Daughter of mother's sister}} & \AR{بِنْت الخَالَة} & \multicolumn{1}{l|}{11 M} & \multirow{2}{*}{16.67 M} \\
        \cline{2-3}
        & \AR{بِنت الخالة} & \multicolumn{1}{l|}{5.67 M} & \\
        \hline
    \end{tabular}
    \label{Table3}
\end{table}

In the contribution collection, a native speaker answers by filling a lexical unit or gap in a row empty slot specified for each concept. Linguistic resources and Google Search are used to provide answers as precise as possible. For example, the \AR{المعاني} Almaany dictionary\footnote{\url{http://www.almaany.com/thesaurus.php}}, Wiktionary\footnote{\url{http://ar.wiktionary.org}}, and the \emph{Fiqh AlArabiyya} typology book \citep{muttaqinfiqh2009} are employed in sequential steps to give a judgment on cousin words in Syrian. Additionally, counting the number of hits returned by the Google search engine is another helpful indicator, where a high count of hits indicates a searching word (i.e., \AR{ابن العمة} meaning ``\textit{son of father's sister},'' has 131.5 million hits) is a lexical unit in Syrian. In contrast, a low count indicates a lexical gap; for example, \AR{الخؤولة} meaning ``\textit{maternal cousin},'' has 158 thousand hits. Google hits of other cousin terms are shown in Table \ref{Table3}. Since Arabic words can be written and read with or without diacritics (i.e., ``\textit{fatha}'' above a letter or ``\textit{kassra}'' under it), thus, each word is typed in two forms.  Note that the content of this matrix cannot be considered the only criterion for gap exploration because word hits may contain a count of other hits resulting from searching in other Arabic dialects for the same word. 

\subsection{Experiment Results}

The overall contribution collection effort resulted in 180~words, 1,108~lexical gaps, and 19~new concepts identified, formalized, and collected. Detailed statistics about the collected gaps and words are shown in Table \ref{Table4}. New concepts were identified in three subdomains: siblings, cousins, and grandchildren. The total number of new concepts, 19, is lower than the sum of new concepts per language due to overlaps across languages: for example, \AR{أَخٌ في الرضاعة} meaning \textit{breastfeeding brother} was found in all seven dialects, \AR{أخت لأم} meaning \textit{maternal sister} was found both in Syrian and in Egyptian, while \AR{أبْيِه} meaning \textit{elder cousin, son of mother's brother} only exists in Egyptian.

Validation was carried out in two phases; in the first phase, words and gaps were validated at the lexicon level by the first author, a Ph.D.~student in lexical semantics and a native speaker of Arabic, and the third author, an Arabic native speaker with linguistic-semantic experience and good knowledge in Arabic dialects. In the second phase, new concepts are verified and approved to be added to the concept layer of the UKC by the second author, a lexical-semantic expert, and the UKC system manager.

Using the lexicon-level validation method, the first author evaluated the collected data in Palestinian and Syrian, while the third author validated the remaining five dialects. Results can be seen in Table~\ref{Table5}, whereby correctness, we understand the number of words (or gaps) validated as correct divided by the total number of words (or gaps). 
In the case of an incorrect word, the validator either provides a correct word or indicates it as a lexical gap. For example, for the Algerian dialect, the correctness of gathered words is~85.71\% and that of gaps is~98.08\%. Four Algerian words were deemed incorrect: \AR{مانّي} for the meaning \textit{maternal grandmother}, \AR{لالّة} for the meaning \textit{paternal grandmother}, \AR{جَدّ} for the meaning \textit{grandfather}, and \AR{باب الشيخ} for the meaning \textit{grandparent}. The validator indicated \textit{maternal grandmother}, \textit{paternal grandmother}, and \textit{grandparent} as gaps, while he replaced the mistaken word \AR{جَدّ} with the correct word \AR{باب الشيخ} for \textit{grandfather}. For gap evaluation, the linguistic expert validates a lexical gap by confirming it as a gap or as a non-gap due to an existing word in a dialect, for which he must provide the correct word. For instance, \textit{Participant 1} identified the meanings \textit{elder sister}, \textit{father's elder sister} and \textit{mother's elder sister} as gaps in Algerian, but the validator did not accept them and provided the polysemous word \AR{لالّة} for each of them. Evidence for validation was obtained from the dictionary \textit{Dictionnaire arabe algérien}\footnote{\url{https://www.lexilogos.com/arabe_algerien.htm}} and from usage attested in Algerian TV films. Upon discussion between the validator and the participants, the mistakes made by the latter can be explained by
misunderstandings of the meanings of certain concepts provided in MSA and English. The validator made sure to exclude or fix the mistakes, bringing the correctness of the final dataset closer to 100\%.

\begin{table}[h!]
    \caption{The count of the diversity items collected and identified in the Arabic dialects. \\}
    \renewcommand{\arraystretch}{1.5}
    \centering
    \begin{tabular}{|l|c|c|c|c|} 
        \hline
        \textbf{Dialects} & \textbf{Words} & \textbf{Gaps w/o new concepts} & \textbf{New concepts} & \textbf{Gaps considering new concepts} \\
        \hline
        Algerian & 28 & 156 & 10 & 165 \\
        \hline
        Egyptian & 32 & 152 & 19 & 152 \\
        \hline
        Moroccan & 22 & 162 & 10 & 169 \\
        \hline
        Palestinian & 23 & 161 & 14 & 166 \\
        \hline
        Syrian & 24 & 160 & 10 & 169 \\
        \hline
        Tunisian & 23 & 161 & 2 & 178 \\
        \hline
        Gulf & 28 & 156 & 14 & 169 \\
        \hline
        \textbf{Total} & \textbf{180} & \textbf{1108} & \textbf{19} & \textbf{1168} \\
        \hline
    \end{tabular}
    \label{Table4}
\end{table}

\begin{table}[h!]
    \caption{Validator evaluation of words and lexical gaps by dialect. \\}
    \renewcommand{\arraystretch}{1.5}
    \centering
    \begin{tabular}{|C{3cm}|C{3cm}C{3cm}|}
        \hline
        \multirow{2}{*}{\textbf{Dialects}} & \multicolumn{2}{c|}{\textbf{Correctness of Native Speaker Contribution}} \\
        \cline{2-3}
        & \multicolumn{1}{C{3cm}|}{\textbf{Words}} & \textbf{Gaps} \\
        \hline
        Algerian & \multicolumn{1}{c|}{85.71\%} & 98.08\% \\
        \hline
        Egyptian & \multicolumn{1}{c|}{96.90\%} & 97.37\% \\
        \hline
        Moroccan & \multicolumn{1}{c|}{95.83\%} & 97.53\% \\
        \hline
        Palestinian & \multicolumn{1}{c|}{100\%} & 98.76\% \\
        \hline
        Syrian & \multicolumn{1}{c|}{91.67\%} & 95.00\% \\
        \hline
        Tunisian & \multicolumn{1}{c|}{95.65\%} & 98.14\% \\
        \hline
        Gulf & \multicolumn{1}{c|}{100\%} & 96.79\% \\
        \hline
        \textbf{Average} & \multicolumn{1}{c|}{\textbf{95.11\%}} & \textbf{97.38\%} \\
        \hline
    \end{tabular}
    \label{Table5}
\end{table}

\begin{figure}[h]
    \centering
    \includegraphics[scale=0.15]{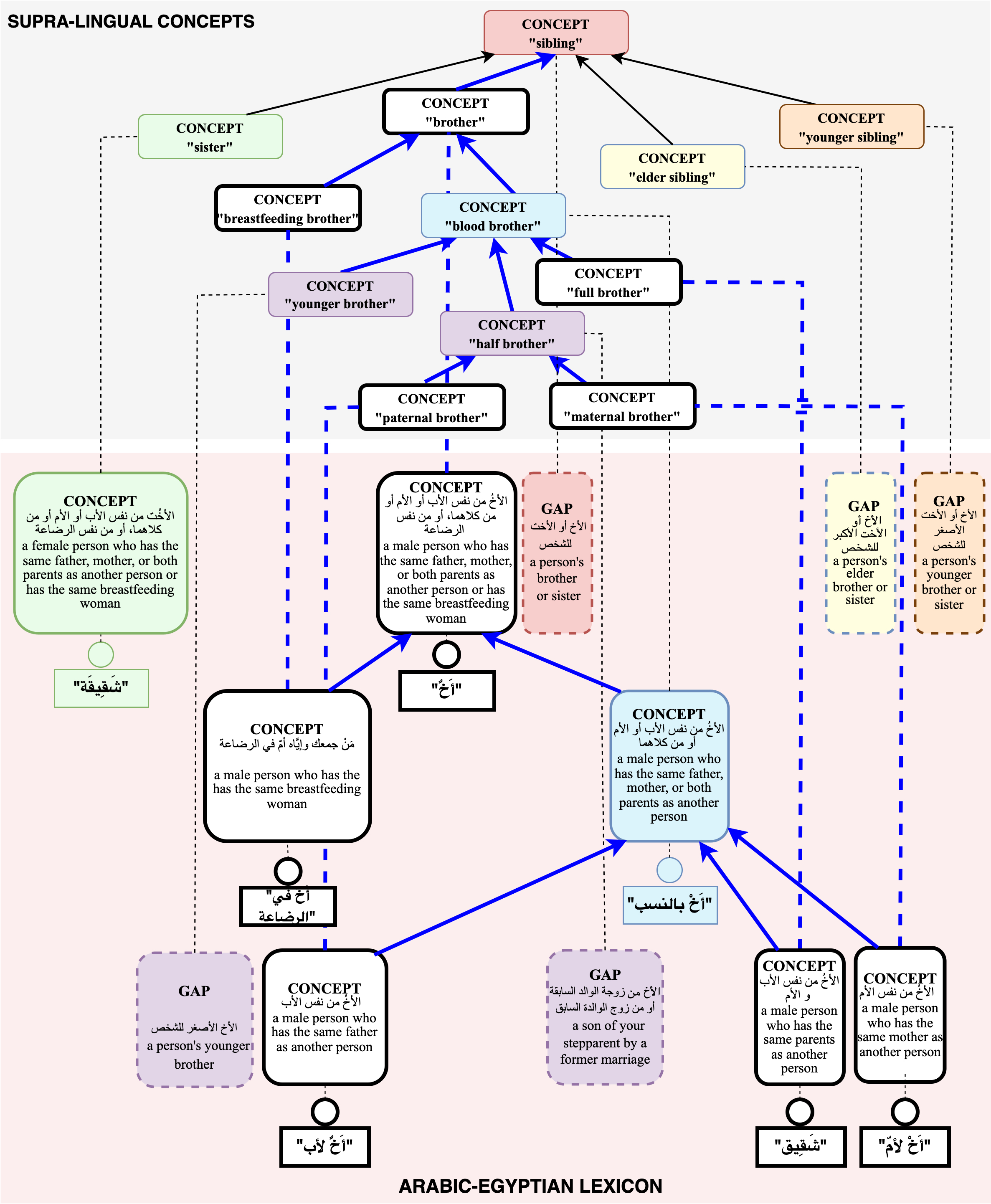}
    \caption{Structural elements in the UKC database after merging new concepts}
    \label{fig5}
\end{figure}

In this study, we use the UKC for creating the input dataset and the domain hierarchy and for storing and visualizing diversity data. Thus, the 19~new concepts were merged with the UKC by reconstructing a domain hierarchy at the supra-lingual concept layer. For example, the hierarchy of siblings was redesigned to contain five new brotherhood concepts and five new sisterhood concepts. For instance, in the Arabic-Egyptian lexicon, as shown in Figure \ref{fig5}, \AR{أَخٌ في الرضاعة} meaning ``\textit{breastfeeding brother},'' is set up as a sub-node for a newly created concept of the brother, ``\textit{a male person who has the same father, mother, or both parents as another person or has the same breastfeeding woman.}'', also, from the figure, can be seen \AR{أخت لأب} meaning ``\textit{paternal brother}'' and \AR{أخت لأم} meaning ``\textit{maternal brother}'' are inserted and connected the half-brother concept. New concepts and lexicalization are marked with white nodes and connected with blue lines.

Additionally, resulting lexical units and gaps were added into UKC lexicons. The website of the UKC provides several services for system users, such as browsable online access to database contents, source materials, and data visualization tools. The interactive exploration of linguistic diversity data in lexicons is the central feature of the website. The user can browse: (1)~all meanings within a language of a word typed in by the user; and (2)~lexicalizations and gaps of a concept in all languages contained in the database. 

\begin{figure}[h]
    \centering
    \includegraphics[scale=0.14]{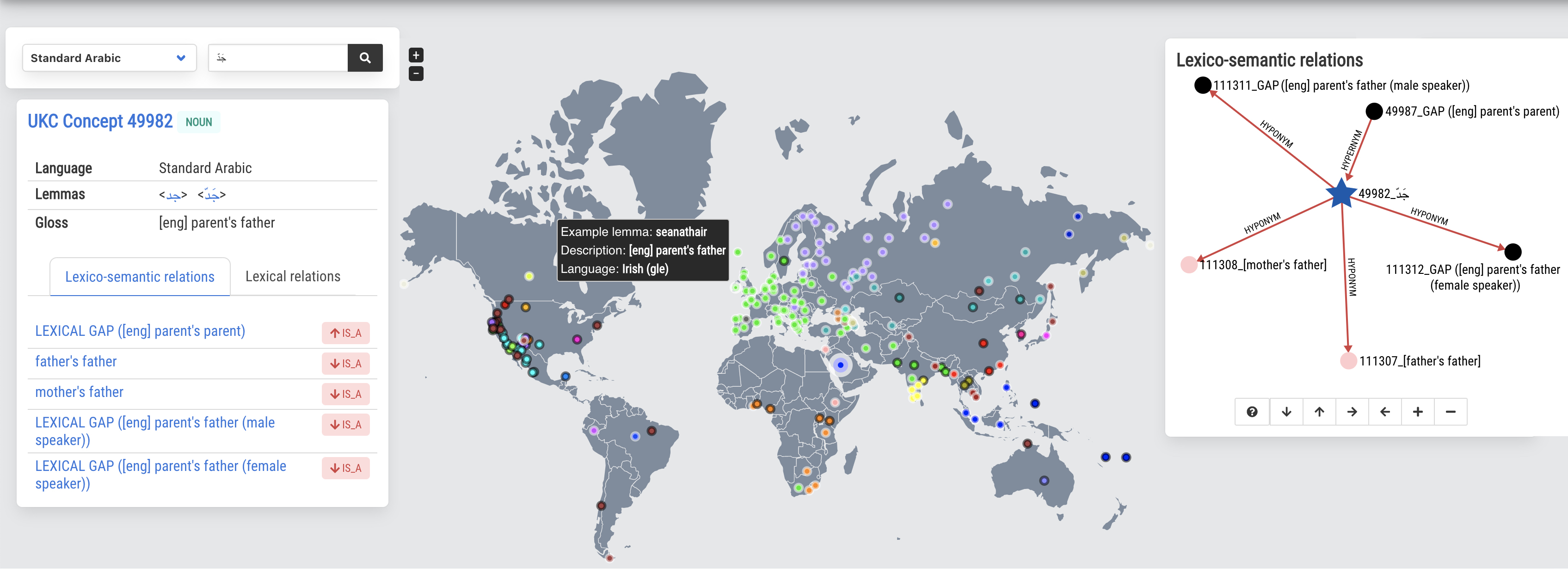}
    \caption{Exploring the concept of \AR{جَدّ} as lexicalized in the Arabic language (left), in the world (middle), and as part of the shared concept hierarchy (right).}
    \label{fig6}
\end{figure}

Figure \ref{fig6} shows a screenshot of the concept exploration functionality describing the concept \AR{جَدّ} meaning ``\textit{parent's father}''. On the left-hand side of the screenshot, details are provided on the lexicalization of the concept in Arabic, such as synonymous words, a definition, and a part of speech. The middle part of the screenshot shows an interactive clickable map of all lexicons that either contain the concept or, on the contrary, lack it due to their languages being known not to lexicalize it. The color-coded dots indicate the language family, while the black circled dot represents a lexical gap. This map presents an instant global typological overview of the concept selected; for instance, from Figure \ref{fig6}, one can see that most languages in Europe lexicalize the concept \AR{جَدّ} while several languages in the American United States do not lexicalize it. Finally, the right-hand side shows the concept \AR{جَدّ} in the context of concept hierarchy, depicted as an interactive graph: the concept, its parent and child concepts, and other lexical-semantic relations (as metonymy and meronymy) are also presented when they exist. Note that the graph only shows a part of the complete hierarchy for usability reasons. Nevertheless, it is navigable and allows the exploration of the whole concept graph in the selected language.

\begin{figure}[h]
    \centering
    \includegraphics[scale=0.5]{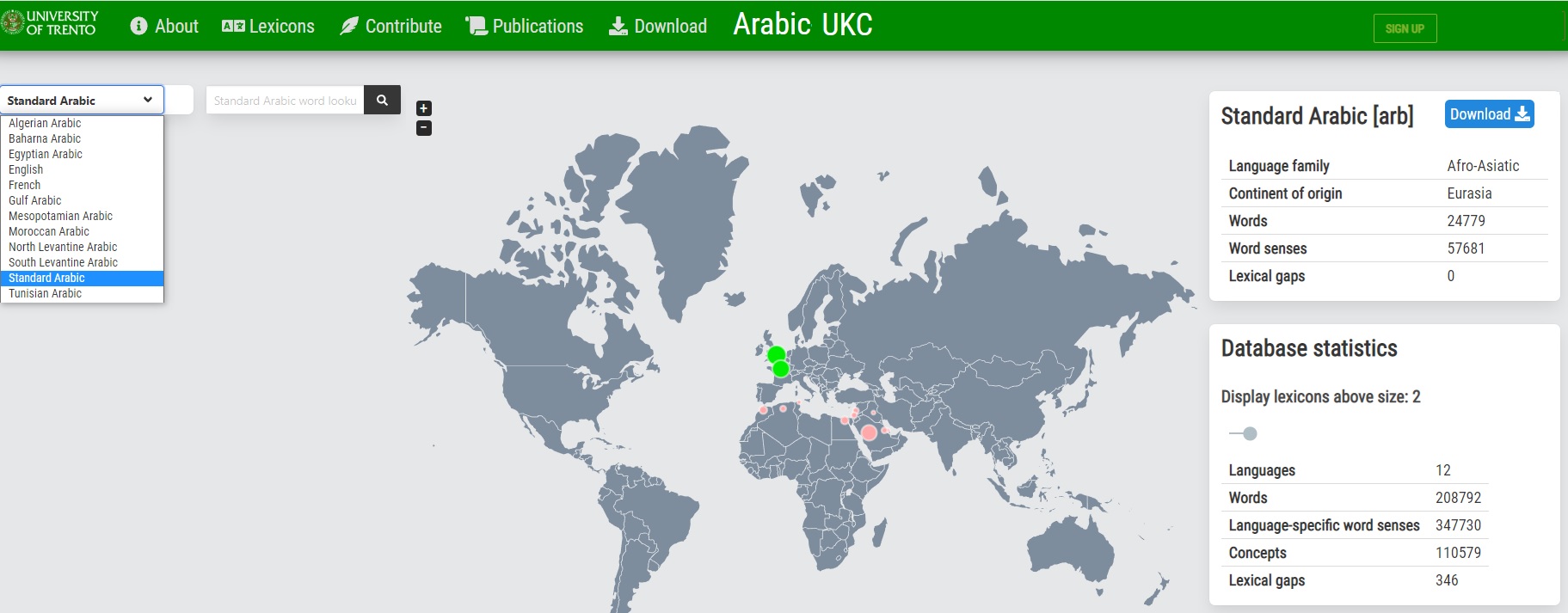}
    \caption{Homepage of the Arabic UKC ongoing project.}
    \label{fig7}
\end{figure}

As mentioned at the beginning of this section, the resulting Arabic dataset will be imported into the Arabic UKC, which is an instance of the UKC system; the top layer contains independent language concepts, and the bottom layer contains twenty lexicons as the number of Arabic dialects. A screenshot of the homepage of the Arabic UKC is shown in Figure \ref{fig7}.

\subsection{Discussion}

The lexical diversity we observed across the seven dialects was higher than our original expectations, with 19~new concepts identified. Ten of these concepts are lexicalised in MSA, such as \AR{أخت في الرضاعة} meaning ``\textit{breastfeeding sister}'' and \AR{أَخٌ لأب} meaning ``\textit{paternal brother}''. The others (nine concepts) are specific to the dialects, such as the Egyptian word \AR{أبْلَه} meaning ``\textit{elder daughter of mother's sister}'', which returns to the Turkish word ``\textit{kuzen}''. Mostly, the origin of these Egyptian-specific concepts is the Ottoman Turkish language, when the Egyptian dialect was influenced by it during the Ottoman occupation of Egypt in the period (1517 AD to 1867 AD).

Several shared meaning overlaps have been found between dialect pairs. Likewise, intersections also existed between gaps. For a given domain $d$ and languages $l_a, ..., l_n$, the formula below calculates the similarity of the two languages in terms of the overlap of lexicalised concepts from that domain, where $\textrm{LexConcepts}(d, l)$ stands for the set of domain concepts that are lexicalized by the language~$l$. 

\begin{equation}
    \textrm{overlap}(d, l_a,...,l_n) = \frac{|\textrm{LexConcepts}(d, l_a)\cap...\cap \textrm{LexConcepts}(d, l_n)|}{\textrm{max}(|\textrm{LexConcepts}(d, l_a)|, ..., |\textrm{LexConcepts}(d, l_n)|)} 
    \label{eq:01}
\end{equation}

\begin{figure}[h]
    \centering
    \includegraphics[scale=0.45]{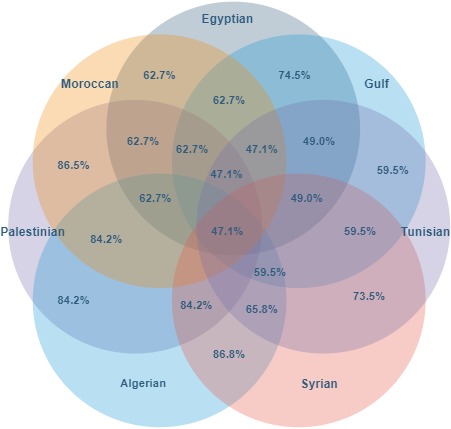}
    \caption{The overlap (percentage of shared lexicalisations) for Arabic dialects.}
    \label{fig8}
\end{figure}

Figure \ref{fig8} shows the overlaps between pairs of Arabic dialects over the kinship domain. For example, the intersection of Egyptian and Gulf dialects gives a shared coverage of 74.5\%, while all dialects are 47.1\% the same.
In the former case, the number of lexicalisations in Egyptian is 51, and in Gulf is 42. Also, 38 of these lexical units are included in both dialects; see the dataset uploaded to GitHub\footnote{\url{https://github.com/HadiPTUK/kinship\_dialect}}. For example, Formula \ref{eq:01} calculates the overlap between Egyptian and Gulf in the Kinship domain ($K$) as follows: 
\[\mathrm{overlap}(K, \mathrm{Egyptian}, \mathrm{Gulf})=\frac{|\mathrm{LexConcepts}(K, \mathrm{Egyptian})\cap \mathrm{LexConcepts}(K, \mathrm{Gulf})|}{\textrm{max}(|\mathrm{LexConcepts}(K, \mathrm{Egyptian})|, |\mathrm{LexConcepts}(K, \mathrm{Gulf})|)}\]
\[\mathrm{overlap}(K, \mathrm{Egyptian}, \mathrm{Gulf})=\frac{38}{\mathrm{max}(51, 42)}=\frac{38}{51}=74.5\%\]
More detail about the analysis of shared coverage between the rest of the Arabic dialects can be found in the same dataset uploaded to GitHub.

We find these overlaps---e.g.~an overlap of 59.5\% between Gulf and Tunisian, or the overall overlap of 47.1\% among all seven dialects---lower than our initial expectations on dialectal variations. Arab dialectologists justify such differences with two major factors: linguistic and religious influence \citep{2014arabic}. By linguistic influence, we refer to the historical interaction of language-speaker communities, which affects the lexicons. Examples are the Egyptian dialect influenced by the Coptic language (historically spoken by the Copts, starting from the third century AD in Roman Egypt) or the Levantine dialect influenced by the Western Aramaic, Canaanite, Turkish, and Greek languages. The Gulf dialect is one of the Peninsular groups, which was influenced by South Arabian Languages. Secondly, the religion of the speaker community also affects the lexicon. Religion is a sociolinguistic variable that shapes how Arabic is spoken. Religion in Arab countries is a matter of group affiliation and is not usually considered an individual choice: one is born a Muslim, Christian, Jew, or Druze, and this becomes a bit like one's ethnicity. So, for example,  within the Egyptian speech community, one can find language mixing between Islamic and Christian terms, and the same in the Levantine community, which consists of a mixing of Muslims, Christians, Jews, and Druze. The Gulf communities, instead, mostly consist of Muslims \citep{al2008arabic}.

\section{Case Study on Diversity Across Indonesian Languages}
\label{secIndonesian}

This section demonstrates the use of the methodology described in Section~\ref{secMethod} on kinship terminology from three Austronesian languages from Indonesia: Indonesian, Javanese, and Banjarese. Contrary to the Arabic dialects in Section~\ref{secArabic}, these three languages are not mutually intelligible.

Indonesia is the fourth most populous country in the world, and it has more than 700 living languages \citep{Eberhard2022}. The national language spoken in Indonesia is Bahasa Indonesia/Indonesian language, which was decided in the historic moment of Youth's Pledge, October 28th, 1928. However, many Indonesians speak more than one language. For example, out of 198~million people that speak Indonesian, 84~million of them speak Javanese \citep{aji2022}. 

Even with the high number of speakers, the count of natural language processing research on Indonesian languages is very low compared to other languages around the world. As of 2020, the count of published papers on the Indonesian language is lower than other languages with less speaker count, such as Polish and Dutch \citep{aji2022}. Not surprisingly, the amount of research on other languages (i.e., Banjarese and Javanese) in Indonesia is much lower than that. It is therefore motivating to conduct this study that discovers the richness of linguistic diversity across three Indonesian languages: standard Indonesian, Banjarese, and Javanese.  In one semantic field, kinship, we have found that diversity is manifested in these languages; for example, in Javanese, the word \textit{ponakan jaler} meaning ``\textit{nephew}'', is a lexical gap in Banjarese, and in the opposite direction, the Banjarese \textit{gulu} meaning ``\textit{parent's second eldest sibling}'' is also a gap in Javanese.

\subsection{Experiment Setup}

As in the Arabic experiment, we use the UKC lexicon to create the
input dataset of kinship terms, which are independent language and formalizing such terms and also new concepts (not existing in the input dataset) identified in this experiment, as shown in the top layer of the UKC in Figure \ref{fig1} for the brotherhood categorization.

In this study, three native speakers (one per language), born and educated (high school level) within the speaker community, were recruited to contribute.
The participants' linguistic backgrounds are listed below:
\begin{enumerate}
    \item \textit{Participant 1}: a native Indonesian speaker with good command of English, Javanese, and Banjarese.
    \item \textit{Participant 2}: a native Banjarese speaker with good command of Indonesian and English.
    \item \textit{Participant 3}: a native Javanese speaker with good command of Indonesian and English.
\end{enumerate}

For each language, an experiment was carried out to identify words and gaps associated with the same 184~kinship concepts as in the Arabic study (see Table~\ref{Table2}). For example, in Banjarese, the dictionary  \emph{Kamus Bahasa Banjar Dialek Hulu-Indonesia} \citep{balai2008} and Google Search hits were used in subsequent steps to provide a precise answer on each concept from the given list of inputs.  Such search steps were also followed by the Banjarese native speaker for the task of judgment on new concepts identified in the uncle/aunt subdomain. For instance, the Banjarese term \emph{gulu}, expressing an uncle/aunt relationship with the meaning of \emph{a parent's second eldest sibling} and attested by the dictionary above, did not previously exist in the UKC or in the KinDiv dataset, nor in \citet{murdock1970}. Indonesian and Javanese native speakers also follow the same steps and use the dictionaries of \citet{indonesia2017} and \citet{javanese2015} for the task of judgment on terms and gaps identified in Indonesian and Javanese, respectively.        

\subsection{Experiment Results}

The overall contribution collection effort resulted in 41~words and 517~lexical gaps. Three new, yet unattested word meanings were also found and formalised as new concepts. All three are used in Banjarese in the uncle/aunt subdomain:
\begin{itemize}
    \item \emph{julak}, meaning \emph{parent's eldest sibling};
    \item \emph{gulu}, meaning \emph{parent's second eldest sibling};
    \item \emph{angah} or \emph{tangah}, meaning \emph{parent's middle elder sibling} (when the number of siblings is odd).
\end{itemize}

Statistics on the data collected for each language are shown in Table~\ref{Table6}.

\begin{table}[h!]
    \caption{The count of the diversity items collected and identified in the Indonesian languages. \\}
    \renewcommand{\arraystretch}{1.5}
    \centering
    \begin{tabular}{|C{2.5cm}|C{2cm}|C{3.5cm}|C{3cm}|C{3.5cm}|}
        \hline
        \textbf{Languages} & \textbf{Words} & \textbf{Gaps w/o new concepts} & \textbf{New concepts} & \textbf{Gaps considering new concepts)} \\
        \hline
        Indonesian & 11 & 173 & 0 & 176 \\
        \hline
        Javanese & 17 & 167 & 0 & 170 \\
        \hline
        Banjarese & 12 & 172 & 3 & 172 \\
        \hline
        \textbf{Total} & 41 & 511 & 3 & 517 \\
        \hline
    \end{tabular}
    \label{Table6}
\end{table}

As in Arabic, a two-step validation was carried out in this study. The first step validated words and gaps contributed by native speakers, carried out by the fourth author, a native Indonesian speaker with a good command of all three languages. The second validation step was done on the concept level, performed by the second author, a lexical-semantic expert and UKC system manager for new concept validation. In this step, the new concepts were verified and approved to be added to the concept layer of the UKC. 

Table \ref{Table7} provides correctness results over native speaker contributions, provided by the validator. Upon discussion between the validator and the contributors, the mistakes made by the latter can be explained by misunderstandings of the meanings of certain concepts, provided in English. The validator made sure to exclude or fix the mistakes, bringing the correctness of the final dataset closer to 100\%.

\begin{table}[h!]
    \caption{Validator evaluation of words and lexical gaps by language. \\}
    \renewcommand{\arraystretch}{1.5}
    \centering
    \begin{tabular}{|C{3cm}|C{3cm}C{3cm}|}
        \hline
        \multirow{2}{*}{\textbf{Languages}} & \multicolumn{2}{c|}{\textbf{Correctness of Native Speaker Contribution}} \\
        \cline{2-3}
        & \multicolumn{1}{C{3cm}|}{\textbf{Words}} & \textbf{Gaps} \\
        \hline
        Indonesian & \multicolumn{1}{c|}{90.91\%} & 98.27\% \\
        \hline
        Javanese & \multicolumn{1}{c|}{94.44\%} & 95.78\% \\
        \hline
        Banjarese & \multicolumn{1}{c|}{91.7\%} & 97.67\% \\
        \hline
        \textbf{Average} & \multicolumn{1}{c|}{\textbf{92.35\%}} & \textbf{97.24\%} \\
        \hline
    \end{tabular}
    \label{Table7}
\end{table}

The produced kinship datasets from this experiment will be merged with the under-construction Indonesian UKC\footnote{\url{http://indonesia.ukc.datascientia.eu/}}, a diversity-aware lexicon for languages spoken in Indonesia, also imported into the main UKC database.

\begin{figure}[h]
    \centering
    \includegraphics[scale=0.14]{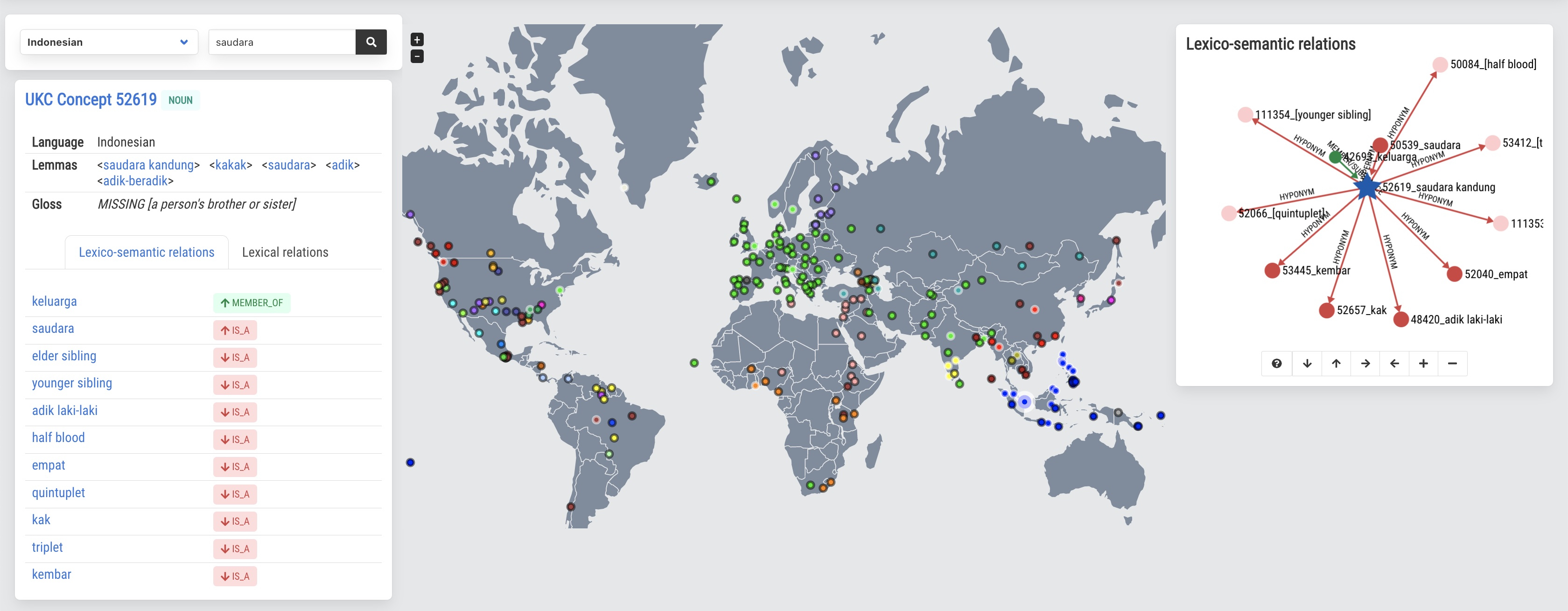}
    \caption{Exploring the concept of \textit{saudara} as lexicalized in the Indonesian language (left), in the world (middle), and as part of the shared concept hierarchy (right).}
    \label{fig9}
\end{figure}

Figure \ref{fig9} shows how UKC explores information about a specific Indonesian word. However, the screenshot provides information about the Indonesian word \textit{saudara}, which means ``\textit{sibling}'' in English. The left-hand side of the screenshot explains synonymous words (lemmas) and the definition of the typed word. The middle of the screenshot displays the map of a global typological overview of the concept. Most languages do not lexicalize this concept, marked by the black-circled dot. Only a few languages lexicalize it, such as Indonesian, Swedish, Ainu, and Malayalam, marked by white-circled dots. The right-hand side shows the lexico-semantic relations of the concept.

\begin{figure}[h]
    \centering
    \includegraphics[scale=1.6]{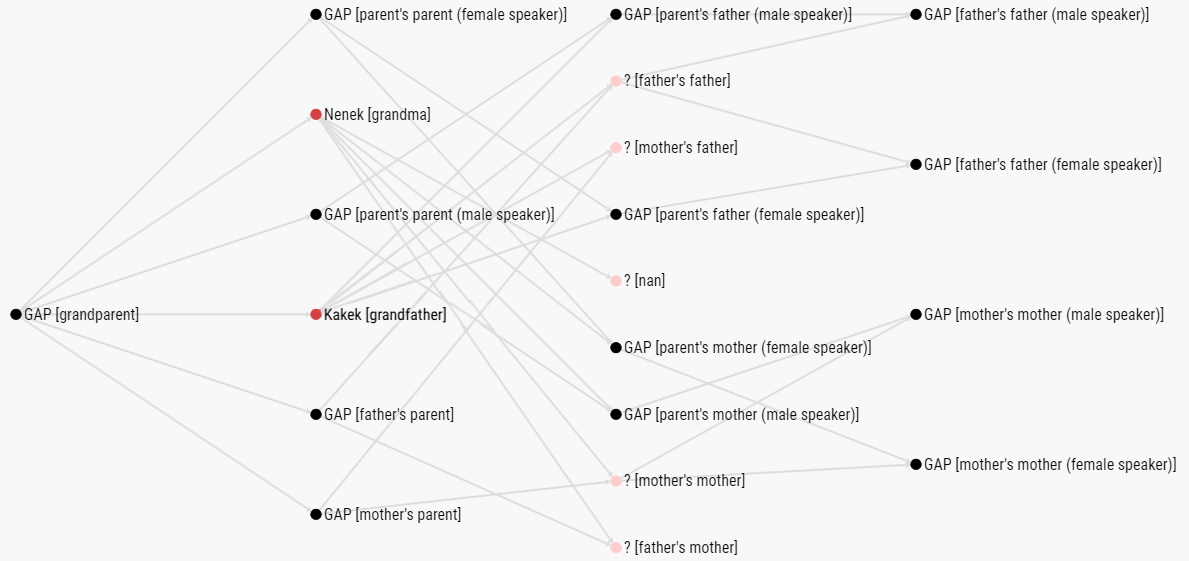}
    \caption{Interactive browser tool showing lexical units and gaps for the grandparent subdomain in Indonesian.}
    \label{fig10}
\end{figure}

The UKC lexicon is also equipped with several interactive visualization services that can be used to browse lexical units and gaps by domain in all supported languages. Figure \ref{fig10} shows an example of using such services in visualizing the content of the grandparent subdomain in Indonesian.

\subsection{Discussion}

\begin{figure}[h]
    \centering
    \includegraphics[scale=0.50]{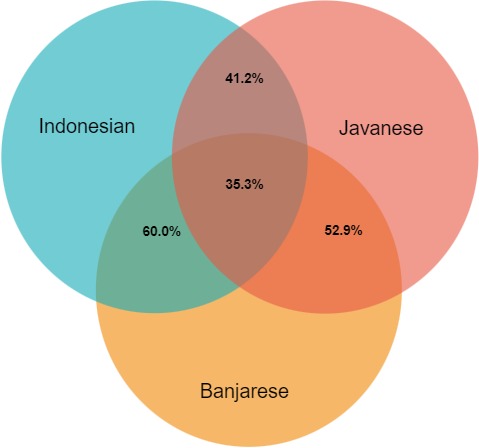}
    \caption{The number of words in the intersection of Indonesian languages according to shared meaning.}
    \label{fig11}
\end{figure}

More than 90\% of our 184~initial kinship concepts were found to be gaps in the three Indonesian languages, as shown in Table~\ref{Table6}. Using Formula \ref{eq:01}, we calculated the overlaps between the Indonesian languages in terms of kinship lexicalisations, shown in Figure~\ref{fig11}. For more details, see the dataset uploaded to the GitHub repository\footnote{\url{https://github.com/HadiPTUK/kinship\_dialect}}. 35.3\% of the concepts are lexicalised by the three Indonesian languages studied. The Javanese--Banjarese  overlap is 52.9\%, Javanese--Indonesian is 60\%, and finally Banjarese--Indonesian is 41.2\%.
Even though all three languages are included in the Malayo-Polynesian branch of the Austronesian language family, Indonesian and Banjarese are considered Malayic languages, while Javanese is not, which is the first reason for this result. Furthermore, these languages exist on different islands in Indonesia; Javanese exists on Java Island, Banjarese is located on the southern part of Borneo Island, and the Indonesian language is based on Malay, which is spoken on Sumatra Island \citep{2003sneddon}, so this geographical barrier restricts interactions between speakers, and each language has developed within its own speech community. 

\begin{figure}[h]
    \centering
    \includegraphics[scale=0.45]{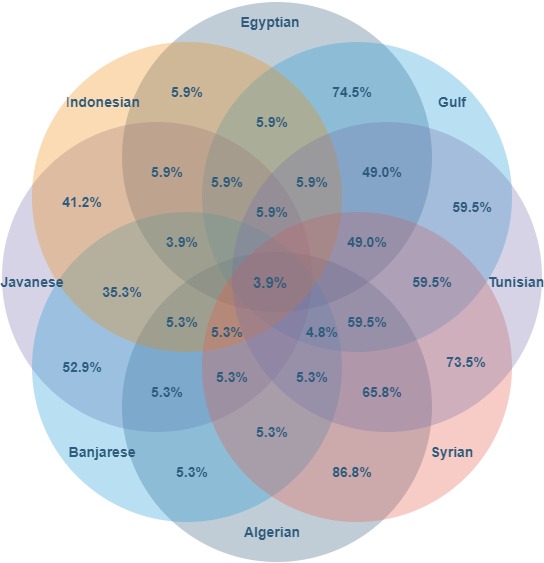}
    \caption{The number of words in the intersection of Indonesian and Arabic languages according to shared meaning.}
    \label{fig12}
\end{figure}

Finally, using Formula \ref{eq:01}, we computed the overlaps between Arabic dialects and Indonesian languages. Figure~\ref{fig12} shows that the ten languages together cover only 3.9\% of the concepts, and the most similar language pair, namely Egyptian--Indonesian, is merely 5.9\%~similar. For researchers in ethnography or comparative linguistics, the observation of such pronounced levels of cross-lingual and cross-cultural diversity may not come as a surprise, as major variations in kin patterns are well known in these domains. On the other hand, we believe that beyond these narrow fields of research, there is a general lack in understanding the depth of diversity in how, through languages, people describe and interpret the world. Most computational linguists and engineers who build language processing systems, as well as the users who trust such systems for their daily activities, do not suspect the breadth of the mental divide across languages that language applications, such as machine translation systems, are meant to bridge. We think that through quantified measures, as we are attempting to do with our simple measure of overlap introduced on p.~\pageref{eq:01}, can be useful to improve our qualitative grasp on diversity, which we consider a promising direction for future research.

Table \ref{Table8} includes statistics of collected words and gaps by domain across Arabic and Indonesian languages. 
The results show that only three words in the domain of cousins are identified in the Indonesian languages, while in Egyptian, 16 words are used around the concept of the cousin.   

\begin{table}[h!]
    \caption{Statistics of the collection of diversity data by domain. \\}
    \renewcommand{\arraystretch}{1.5}
    \centering
    \begin{tabular}{|C{3cm}|C{3cm}|C{3cm}|}
        \hline
        \textbf{Domains} & \textbf{Words} & \textbf{Gaps} \\
        \hline
        Grandparents & 21 & 169 \\
        \hline
        Grandchildren & 19 & 251 \\
        \hline
        Siblings & 37 & 173 \\
        \hline
        Uncle/Aunt & 44 & 226 \\
        \hline
        Nephew/Niece & 33 & 297 \\
        \hline
        Cousins & 67 & 503 \\
        \hline
        \textbf{Total} & 221 & 1619 \\
        \hline
    \end{tabular}
    \label{Table8}
\end{table}

\section{Related Work}
\label{secRelatedWork}

Ethnologists and linguists have for a long time studied how family structures map to kinship terminology across languages and social groups. The most famous and comprehensive ethnographic study on kin term patterns is that of \citet{murdock1970}, upon which our work also indirectly relied: our cross-lingual formalisation of kin terms is based on the one provided by the KinDiv resource, itself in part derived from Murdock's data. KinDiv covers 699~languages and is a computer-processable database that can also be exploited for applications in computational linguistics. Our results provide linguistic evidence in seven Arabic dialects and three Indonesian languages that do not figure in these resources.

The exploration of kin terminology and the building of large-scale databases on the topic has also been the subject of more recent efforts---we only cite two examples here. The AustKin project\footnote{\url{http://austkin.net}} has produced a large-scale database on kin terms in hundreds of indigenous Australian languages. The recent Kinbank database \citep{kinbank2023} is a comprehensive resource on kinship terminology, covering over 1,173~languages, with a broad coverage of kinship subdomains. As Kinbank was released after the initial submission of our paper, we did not rely on it for our work. We consider our research as complementary to Kinbank: concentrating on a relatively low number of dialects and languages, our results could, in principle, be integrated into Kinbank in order to extend its coverage. And vice versa, we see potential in using Kinbank data in order to cross-validate and possibly to extend the Indonesian terms we collected (as the three Indonesian languages of our study are also covered by Kinbank). There is, however, an important methodological difference between the our and Kinbank's way of representing terms: Kinbank does not explicitly indicate lexical gaps. For example, our work considers the concept of \textit{son of father's brother as pronounced by a male speaker} to be a lexical gap in Javanese, while Kinbank maps the Javanese term \emph{sedulur misan}, simply meaning \emph{cousin}, to this and 95~other meanings. Our work, instead, identifies the Javanese term as the general meaning of \emph{cousin} and considers all other (more specific) cousin terms as lexical gaps. This distinction is useful in comparative linguistics and cross-lingual applications where the explicit indication of the lack of precise meaning equivalence can be exploited.

Concepticon \citep{list2016concepticon} is `a resource for linking concept lists' frequently used in comparative linguistics. The \emph{concept sets} of Concepticon serve the same purpose as the supra-lingual concepts of the UKC in our study, namely to provide meaning-based mappings among lists of terms (aka \emph{concept lists} in Concepticon) across languages. As of mid-2023, Concepticon consists of nearly 4,000 concept sets, principally targeting core vocabularies (basic-level categories) that are the main subject of study of historical and comparative linguistics. Concepticon is under continuous development and has more recently evolved from a flat list of meanings to a hierarchy with broader--narrower relations. At the time of writing, the kinship domain seems to be partially represented in Concepticon: while sibling or grandparent relations are widely covered, fine-grained cousin relationships are mostly missing from it. The UKC, which contains over 100,000 supra-lingual concepts and a wide range of lexical and lexico-semantic relations, was a more suitable resource for our study due to its more complete coverage of the kinship domain and its explicit support for representing term untranslatability via lexical gaps. 

Multilingual computational applications being in the core of our focus, we also review relevant resources from computational linguistics. For NLP applications, the most popular and widely-known representation of lexico-semantic knowledge is that of \emph{wordnets} that follow the general structure of the original English \emph{Princeton WordNet} \citep{miller1995}.
The \textit{wordnet expansion} approach by \citet{fellbaum2012}---an expert-driven lexicon translation effort---is frequently used to produce new wordnets for lower-resourced languages: this approach consists of `translating' (i.e.~finding lexicalizations for) English WordNet concepts (`synsets' in wordnet terminology) into the target language.
While this is a straightforward approach that produces resources that remain cross-lingually linked, its downside is that the translation approach cannot involve concepts and words specific to the target language and not present in the source language (which in most cases is English). In cases of diverse conceptualisations of the world, the translation approach often results in incorrect approximations. To take the example of Arabic, both versions of the Arabic Wordnet \citep{elkateb2006,abouenour2013} map the English synset of \emph{uncle} (``\textit{the brother of your father or mother; the husband of your aunt}'')  to the Arabic synset of \AR{عم}, which means ``\textit{the brother of your father}.''

A similar situation is observed for Indonesian. As far as we know, the only Indonesian Wordnet currently accessible is Bahasa Wordnet---a bilingual Wordnet for standard Indonesian and Malay languages \citep{noor2011}. It was formed by merging three different wordnets (one in Indonesian and two in Malay) developed mainly by the same expansion approach from PWN. Due to this approach, many English words that have no equivalents in Indonesian are incorrectly mapped, resulting in meaning loss. For example, in Bahasa Wordnet, the English word \emph{sister}, which means ``\textit{a female person who has the same parents as another person},'' was mapped to the Indonesian word \emph{kakak} which means ``\textit{elder sibling}.''

Finally, we mention MultiWordNet as an early effort at improving the representation of linguistic diversity in multilingual lexical databases \citep{pianta2002}. It is a multilingual lexicon that was built using the \textit{merge} method that, contrary to the translation-based expand approach presented above, maps together existing high-quality bilingual dictionaries. MultiWordNet explicitly represents lexical gaps in its Italian and Hebrew wordnets: about 1,000 in Italian and about 300 in Hebrew \citep{bentivogli2000,ordan2007}. MultiWordNet, however, is a discontinued effort that does not cover the kinship domain and is thus was not suitable for our purposes.

The methodology we present in Section~\ref{secMethod} follows neither the expansion nor the merge approach but a third one, more adapted to diversity-aware lexicography: our starting point is a supra-lingual, diversity-aware conceptualization of the domain of study (kinship in our case). The task of \emph{contribution collection} is performed by native speakers with respect to the supra-lingual concept hierarchy based on evidence from comparative linguistics and covering a wide range of languages. While there is no guarantee that our initial conceptualization is complete---indeed, it was not the case in our study---it is less biased towards the concepts of a single language and speaker community than the expansion approach.

\section{Conclusions and Future Work}
\label{secConclusion}

Our paper formally captures lexical diversity across languages and dialects by representing language- or dialect-specific concepts and linguistic gaps. It introduces a systematic method to produce such data in a human-based manner from one semantic domain rather than from general domains, as the efforts of covering the WordNet domains \citep{magnini2000} that have been conducted in building these wordnets, Mongolian \citep{batsuren2019}, Unified Scottish Gaelic\citep{bella2020}, and MultiWordNet\citep{pianta2002}.

The method is verified through two large-scale case studies on kinship terminology, a domain known to be diverse across languages and cultures: one case study deals with seven Arabic dialects, while the other one with three Indonesian languages. The experiments show that our method outperforms the existing methods in terms of the quantity of explored gaps and words and the quality of results. Overall efforts resulted in 1619 gaps, and 223 words were identified in 10 languages and dialects. Moreover, 22 new word meanings with respect to the imported list of independent-language concepts from the UKC are explored in this research.

In future work, we plan to automate the method presented in this paper and apply it to new languages, such as the rest of the Arabic dialects and Indonesian language, as well as to new domains that are known to be diverse, such as body parts, food, color, or visual objects \citep{giunchiglia2021, giunchiglia2023}. 

Finally, diversity-aware lexicons such as the UKC (which includes our produced datasets) provide essential information to cross-lingual applications, such as multilingual NLP tasks or cross-lingual language models. In the future, we plan to use this resource in implementing one such application, i.e., machine translation.

\section*{Conflict of Interest Statement}

The authors declare no conflict of interest.

\section*{Author Contributions}
FG and GB conceptualized and supervised the study. GB and HK imported and formatted the dataset of inputs. HK wrote the original manuscript draft and performed the Arabic experiments. AF and HK validated the collected Arabic data at the lexicon level. SD performed the Indonesian experiments and validated the results at the lexicon level. GB validated the identified diverse data at the concept level. FG, GB, AF, and HK analyzed the Arabic and Indonesian data. FG, GB, AF, SD, and HK reviewed and edited the manuscript. All authors contributed to the research and approved the submitted version.

\section*{Acknowledgments}
We thank the University of Trento and Palestine Technical University—Kadoori for their support.

\section*{Data Availability Statement}
The diversity-aware datasets of the kinship generated and analyzed for this study can be found in the GitHub repository (\textit{https://github.com/HadiPTUK/kinship\_dialect}).

\bibliographystyle{apalike}
\bibliography{main}  %%% Uncomment this line and comment out the ``thebibliography'' section below to use the external .bib file (using bibtex) .

\begin{thebibliography}{}

\bibitem[Abouenour et~al., 2013]{abouenour2013}
Abouenour, L., Bouzoubaa, K., and Rosso, P. (2013).
\newblock On the evaluation and improvement of {Arabic WordNet} coverage and usability.
\newblock {\em Language Resources and Evaluation}, 47:891--917.

\bibitem[Aji et~al., 2022]{aji2022}
Aji, A.~F., Winata, G.~I., Koto, F., Cahyawijaya, S., Romadhony, A., Mahendra, R., Kurniawan, K., Moeljadi, D., Prasojo, R.~E., Baldwin, T., Lau, J.~H., and Ruder, S. (2022).
\newblock One country, 700+ languages: {NLP} challenges for underrepresented languages and dialects in {I}ndonesia.
\newblock In {\em Proceedings of the 60th Annual Meeting of the Association for Computational Linguistics (Volume 1: Long Papers)}, pages 7226--7249. Association for Computational Linguistics.

\bibitem[Al-Wer, 2008]{al2008arabic}
Al-Wer, E. (2008).
\newblock {A}rabic languages, variation in.
\newblock In Brown, K. and Ogilvie, S., editors, {\em Concise Encyclopedia of Languages of the World}, pages 53--56. Elsevier Ltd., Oxford.

\bibitem[Anderson et~al., 2018]{anderson2018}
Anderson, C., Tresoldi, T., Chacon, T., Fehn, A.-M., Walworth, M., Forkel, R., and List, J.-M. (2018).
\newblock A cross-linguistic database of phonetic transcription systems.
\newblock In {\em Yearbook of the Poznan Linguistic Meeting}, volume~4, pages 21--53. De Gruyter Open.

\bibitem[Arora et~al., 2021]{arora2021}
Arora, A., Farris, A., Gopalakrishnan, R., and Basu, S. (2021).
\newblock Bh\=a\d{s}\=acitra: Visualising the dialect geography of {S}outh {A}sia.
\newblock In {\em Proceedings of the 2nd International Workshop on Computational Approaches to Historical Language Change 2021}, pages 51--57. Association for Computational Linguistics.

\bibitem[{Badan Pengembangan dan Pembinaan Bahasa}, 2017]{indonesia2017}
{Badan Pengembangan dan Pembinaan Bahasa} (2017).
\newblock {\em Kamus Besar Bahasa Indonesia}.
\newblock Badan Pengembangan dan Pembinaan Bahasa, Kementerian Pendidikan dan Kebudayaan, Indonesia.

\bibitem[{Balai Bahasa Banjarmasin}, 2008]{balai2008}
{Balai Bahasa Banjarmasin} (2008).
\newblock {\em Kamus Bahasa Banjar Dialek Hulu-Indonesia}.
\newblock Departemen Pendidikan Nasional, Pusat Bahasa, Balai Bahasa Banjarmasin, Indonesia.

\bibitem[Batsuren et~al., 2019]{batsuren2019}
Batsuren, K., Bella, G., and Giunchiglia, F. (2019).
\newblock {C}og{N}et: A large-scale cognate database.
\newblock In {\em Proceedings of the 57th Annual Meeting of the Association for Computational Linguistics}, pages 3136--3145. Association for Computational Linguistics.

\bibitem[Batsuren et~al., 2022]{batsuren2022}
Batsuren, K., Goldman, O., Khalifa, S., Habash, N., Kiera{\'s}, W., Bella, G., Leonard, B., Nicolai, G., Gorman, K., Ate, Y.~G., Ryskina, M., Mielke, S., Budianskaya, E., El-Khaissi, C., Pimentel, T., Gasser, M., Lane, W.~A., Raj, M., Coler, M., Samame, J. R.~M., Camaiteri, D.~S., Rojas, E.~Z., L{\'o}pez~Francis, D., Oncevay, A., L{\'o}pez~Bautista, J., Villegas, G. C.~S., Hennigen, L.~T., Ek, A., Guriel, D., Dirix, P., Bernardy, J.-P., Scherbakov, A., Bayyr-ool, A., Anastasopoulos, A., Zariquiey, R., Sheifer, K., Ganieva, S., Cruz, H., Karah{\'o}{\v{g}}a, R., Markantonatou, S., Pavlidis, G., Plugaryov, M., Klyachko, E., Salehi, A., Angulo, C., Baxi, J., Krizhanovsky, A., Krizhanovskaya, N., Salesky, E., Vania, C., Ivanova, S., White, J., Maudslay, R.~H., Valvoda, J., Zmigrod, R., Czarnowska, P., Nikkarinen, I., Salchak, A., Bhatt, B., Straughn, C., Liu, Z., Washington, J.~N., Pinter, Y., Ataman, D., Wolinski, M., Suhardijanto, T., Yablonskaya, A., Stoehr, N., Dolatian, H., Nuriah, Z., Ratan, S., Tyers,
  F.~M., Ponti, E.~M., Aiton, G., Arora, A., Hatcher, R.~J., Kumar, R., Young, J., Rodionova, D., Yemelina, A., Andrushko, T., Marchenko, I., Mashkovtseva, P., Serova, A., Prud{'}hommeaux, E., Nepomniashchaya, M., Giunchiglia, F., Chodroff, E., Hulden, M., Silfverberg, M., McCarthy, A.~D., Yarowsky, D., Cotterell, R., Tsarfaty, R., and Vylomova, E. (2022).
\newblock {U}ni{M}orph 4.0: {U}niversal {M}orphology.
\newblock In {\em Proceedings of the Thirteenth Language Resources and Evaluation Conference}, pages 840--855, Marseille, France. European Language Resources Association.

\bibitem[Bella et~al., 2022a]{bella2022b}
Bella, G., Batsuren, K., Khishigsuren, T., and Giunchiglia, F. (2022a).
\newblock Linguistic diversity and bias in online dictionaries.
\newblock In Lena, K., editor, {\em Frontiers in African Digital Research}, pages 173--186. Institute of African Studies.

\bibitem[Bella et~al., 2022b]{bella2022a}
Bella, G., Byambadorj, E., Chandrashekar, Y., Batsuren, K., Cheema, D., and Giunchiglia, F. (2022b).
\newblock Language diversity: Visible to humans, exploitable by machines.
\newblock In {\em Proceedings of the 60th Annual Meeting of the Association for Computational Linguistics: System Demonstrations}, pages 156--165. Association for Computational Linguistics.

\bibitem[Bella et~al., 2023]{bella2023towards}
Bella, G., Helm, P., Koch, G., and Giunchiglia, F. (2023).
\newblock Towards bridging the digital language divide.
\newblock {\em arXiv preprint arXiv:2307.13405}.

\bibitem[Bella et~al., 2020]{bella2020}
Bella, G., McNeill, F., Gorman, R., Donna{\'\i}le, C.~{\'O}., MacDonald, K., Chandrashekar, Y., Freihat, A.~A., and Giunchiglia, F. (2020).
\newblock A major {W}ordnet for a minority language: {S}cottish {G}aelic.
\newblock In {\em Proceedings of the Twelfth Language Resources and Evaluation Conference}, pages 2812--2818. European Language Resources Association.

\bibitem[Bentivogli and Pianta, 2000]{bentivogli2000}
Bentivogli, L. and Pianta, E. (2000).
\newblock Looking for lexical gaps.
\newblock In Heid, U. and Evert, S., editors, {\em Proceedings of the 9th EURALEX International Congress}, pages 663--669. Institut für Maschinelle Sprachverarbeitung.

\bibitem[Carling et~al., 2018]{carling2018}
Carling, G., Larsson, F., Cathcart, C.~A., Johansson, N., Holmer, A., Round, E., and Verhoeven, R. (2018).
\newblock {Diachronic Atlas of Comparative Linguistics (DiACL)}—a database for ancient language typology.
\newblock {\em PLOS ONE}, 13(10):e0205313.

\bibitem[Catford, 1965]{catford1965}
Catford, J.~C. (1965).
\newblock {\em A Linguistic Theory of Translation}.
\newblock Oxford University Press London, London.

\bibitem[Dryer and Haspelmath, 2013]{dryer2013}
Dryer, M.~S. and Haspelmath, M., editors (2013).
\newblock {\em WALS Online (v2020.3)}.
\newblock Zenodo.

\bibitem[Eberhard et~al., 2022]{Eberhard2022}
Eberhard, D., Simons, G.~F., and Fenning, C.~D. (2022).
\newblock {\em Ethnologue: Languages of Africa and Europe}.
\newblock SIL International Publications.

\bibitem[Elkateb et~al., 2006]{elkateb2006}
Elkateb, S., Black, W., Rodr{\'\i}guez, H., Alkhalifa, M., Vossen, P., Pease, A., and Fellbaum, C. (2006).
\newblock Building a {W}ord{N}et for {A}rabic.
\newblock In {\em Proceedings of the Fifth International Conference on Language Resources and Evaluation ({LREC}{'}06)}, pages 29--34. European Language Resources Association.

\bibitem[Fellbaum and Vossen, 2012]{fellbaum2012}
Fellbaum, C. and Vossen, P. (2012).
\newblock Challenges for a multilingual wordnet.
\newblock {\em Language Resources and Evaluation}, 46:313--326.

\bibitem[Georgakopoulos et~al., 2022]{georgakopoulos2022}
Georgakopoulos, T., Grossman, E., Nikolaev, D., and Polis, S. (2022).
\newblock Universal and macro-areal patterns in the lexicon: A case-study in the perception-cognition domain.
\newblock {\em Linguistic Typology}, 26(2):439--487.

\bibitem[Giunchiglia and Bagchi, 2021]{giunchiglia2021}
Giunchiglia, F. and Bagchi, M. (2021).
\newblock Classifying concepts via visual properties.
\newblock {\em arXiv preprint arXiv:2105.09422}.

\bibitem[Giunchiglia et~al., 2023]{giunchiglia2023}
Giunchiglia, F., Bagchi, M., and Diao, X. (2023).
\newblock A semantics-driven methodology for high-quality image annotation.
\newblock {\em arXiv preprint arXiv:2307.14119}.

\bibitem[Giunchiglia et~al., 2017]{giunchiglia2017}
Giunchiglia, F., Batsuren, K., and Bella, G. (2017).
\newblock Understanding and exploiting language diversity.
\newblock In {\em Proceedings of the Twenty-Sixth International Joint Conference on Artificial Intelligence, {IJCAI-17}}, pages 4009--4017.

\bibitem[Giunchiglia et~al., 2018]{giunchiglia2018}
Giunchiglia, F., Batsuren, K., and Freihat, A.~A. (2018).
\newblock One world--seven thousand languages.
\newblock In Gelbukh, A., editor, {\em Proceedings 19th International Conference on Computational Linguistics and Intelligent Text Processing, CiCling2018}, pages 18--24. Springer.

\bibitem[Helm et~al., 2023]{helm2023diversity}
Helm, P., Bella, G., Koch, G., and Giunchiglia, F. (2023).
\newblock Diversity and language technology: How techno-linguistic bias can cause epistemic injustice.
\newblock {\em arXiv preprint arXiv:2307.13714}.

\bibitem[Kay and Cook, 2016]{kay2016}
Kay, P. and Cook, R.~S. (2016).
\newblock World color survey.
\newblock In Luo, M.~R., editor, {\em Encyclopedia of Color Science and Technology}, pages 1265--1271. Springer, New York.

\bibitem[Kemp and Regier, 2012]{kemp2012}
Kemp, C. and Regier, T. (2012).
\newblock Kinship categories across languages reflect general communicative principles.
\newblock {\em Science}, 336(6084):1049--1054.

\bibitem[Khishigsuren et~al., 2022]{Khishigsuren2022}
Khishigsuren, T., Bella, G., Batsuren, K., Freihat, A.~A., Chandran~Nair, N., Ganbold, A., Khalilia, H., Chandrashekar, Y., and Giunchiglia, F. (2022).
\newblock Using linguistic typology to enrich multilingual lexicons: the case of lexical gaps in kinship.
\newblock In {\em Proceedings of the Thirteenth Language Resources and Evaluation Conference}, pages 2798--2807, Marseille, France. European Language Resources Association.

\bibitem[Kirby et~al., 2016]{kirby2016}
Kirby, K.~R., Gray, R.~D., Greenhill, S.~J., Jordan, F.~M., Gomes-Ng, S., Bibiko, H.-J., Blasi, D.~E., Botero, C.~A., Bowern, C., Ember, C.~R., et~al. (2016).
\newblock {D-PLACE}: A global database of cultural, linguistic and environmental diversity.
\newblock {\em PLOS ONE}, 11(7):e0158391.

\bibitem[Kopecka and Narasimhan, 2012]{kopecka2012}
Kopecka, A. and Narasimhan, B. (2012).
\newblock {\em Events of putting and taking: A crosslinguistic perspective}.
\newblock John Benjamins Publishing.

\bibitem[Lehrer, 1970]{lehrer1970}
Lehrer, A. (1970).
\newblock Notes on lexical gaps.
\newblock {\em Journal of Linguistics}, 6(2):257--261.

\bibitem[Levinson and Wilkins, 2006]{levinson2006}
Levinson, S.~C. and Wilkins, D.~P. (2006).
\newblock {\em Grammars of Space: Explorations in Cognitive Diversity}.
\newblock Cambridge University Press, Cambridge.

\bibitem[List et~al., 2016]{list2016concepticon}
List, J.-M., Cysouw, M., and Forkel, R. (2016).
\newblock {C}oncepticon: A resource for the linking of concept lists.
\newblock In {\em Proceedings of the Tenth International Conference on Language Resources and Evaluation ({LREC}'16)}, pages 2393--2400. European Language Resources Association.

\bibitem[Magnini and Cavagli{\`a}, 2000]{magnini2000}
Magnini, B. and Cavagli{\`a}, G. (2000).
\newblock Integrating subject field codes into {W}ord{N}et.
\newblock In {\em Proceedings of the Second International Conference on Language Resources and Evaluation ({LREC}{'}00)}. European Language Resources Association.

\bibitem[Majid et~al., 2007]{majid2007}
Majid, A., Bowerman, M., van Staden, M., and Boster, J.~S. (2007).
\newblock The semantic categories of cutting and breaking events: A crosslinguistic perspective.
\newblock {\em Cognitive Linguistics}, 18(2):133--152.

\bibitem[McCarthy et~al., 2019]{mccarthy2019}
McCarthy, A.~D., Wu, W., Mueller, A., Watson, B., and Yarowsky, D. (2019).
\newblock Modeling color terminology across thousands of languages.
\newblock In {\em Proceedings of the 2019 Conference on Empirical Methods in Natural Language Processing and the 9th International Joint Conference on Natural Language Processing (EMNLP-IJCNLP)}, pages 2241--2250. Association for Computational Linguistics.

\bibitem[Miller, 1995]{miller1995}
Miller, G.~A. (1995).
\newblock {W}ord{N}et: A lexical database for {E}nglish.
\newblock {\em Communications of the ACM}, 38(11):39--41.

\bibitem[Murdock, 1970]{murdock1970}
Murdock, G.~P. (1970).
\newblock Kin term patterns and their distribution.
\newblock {\em Ethnology}, 9(2):165--208.

\bibitem[Muttaqin, 2009]{muttaqinfiqh2009}
Muttaqin, Z. (2009).
\newblock Fiqh lughah dalam literatur {A}rab klasik.
\newblock {\em Afaq 'Arabiyah: Jurnal Kebahasaaraban dan Pendidikan Bahasa Arab}, 4(2):107--122.

\bibitem[Noor et~al., 2011]{noor2011}
Noor, N. H. B.~M., Sapuan, S., and Bond, F. (2011).
\newblock Creating the open {W}ordnet {B}ahasa.
\newblock In {\em Proceedings of the 25th Pacific Asia Conference on Language, Information and Computation}, pages 255--264. Institute of Digital Enhancement of Cognitive Processing, Waseda University.

\bibitem[Ordan and Wintner, 2007]{ordan2007}
Ordan, N. and Wintner, S. (2007).
\newblock {H}ebrew {W}ord{N}et: a test case of aligning lexical databases across languages.
\newblock {\em International Journal of Translation}, 19(1):39--58.

\bibitem[Passmore et~al., 2023]{kinbank2023}
Passmore, S., Barth, W., Greenhill, S.~J., Quinn, K., Sheard, C., Argyriou, P., Birchall, J., Bowern, C., Calladine, J., Deb, A., Diederen, A., Metsäranta, N.~P., Araujo, L.~H., Schembri, R., Hickey-Hall, J., Honkola, T., Mitchell, A., Poole, L., R\'{a}cz, P.~M., Roberts, S.~G., Ross, R.~M., Thomas-Colquhoun, E., Evans, N., and Jordan, F.~M. (2023).
\newblock {K}inbank: A global database of kinship terminology.
\newblock {\em PLOS ONE}, 18(5):e0283218.

\bibitem[Pianta et~al., 2002]{pianta2002}
Pianta, E., Bentivogli, L., and Girardi, C. (2002).
\newblock Developing an aligned multilingual database.
\newblock In {\em Proceedings of the 1st International WordNet Conference}, pages 293--302. Global Wordnet Association.

\bibitem[Plungyan, 2011]{plungyan2011}
Plungyan, V. (2011).
\newblock Modern linguistic typology.
\newblock {\em Herald of the Russian Academy of Sciences}, 81(2):101--113.

\bibitem[Reznikova et~al., 2012]{reznikova2012}
Reznikova, T., Rakhilina, E., and Bonch-Osmolovskaya, A. (2012).
\newblock Towards a typology of pain predicates.
\newblock {\em Linguistics}, 50(3):421--465.

\bibitem[Roberson et~al., 2005]{roberson2005}
Roberson, D., Davidoff, J., Davies, I.~R., and Shapiro, L.~R. (2005).
\newblock Color categories: Evidence for the cultural relativity hypothesis.
\newblock {\em Cognitive Psychology}, 50(4):378--411.

\bibitem[Rzymski et~al., 2020]{rzymski2020}
Rzymski, C., Tresoldi, T., Greenhill, S.~J., Wu, M.-S., Schweikhard, N.~E., Koptjevskaja-Tamm, M., Gast, V., Bodt, T.~A., Hantgan, A., Kaiping, G.~A., Chang, S., Lai, Y., Morozova, N., Arjava, H., Hübler, N., Koile, E., Pepper, S., Proos, M., Epps, B.~V., Blanco, I., Hundt, C., Monakhov, S., Pianykh, K., Ramesh, S., Gray, R.~D., Forkel, R., and List, J.-M. (2020).
\newblock The database of cross-linguistic colexifications, reproducible analysis of cross-linguistic polysemies.
\newblock {\em Scientific Data}, 7(1):1--13.

\bibitem[Salesky et~al., 2020]{salesky2020}
Salesky, E., Chodroff, E., Pimentel, T., Wiesner, M., Cotterell, R., Black, A.~W., and Eisner, J. (2020).
\newblock A corpus for large-scale phonetic typology.
\newblock In {\em Proceedings of the 58th Annual Meeting of the Association for Computational Linguistics}, pages 4526--4546. Association for Computational Linguistics.

\bibitem[Sneddon, 2003]{2003sneddon}
Sneddon, J. (2003).
\newblock {\em The Indonesian Language}.
\newblock University of New South Wales Press Ltd, Sydney.

\bibitem[Utomo, 2015]{javanese2015}
Utomo, S.~S. (2015).
\newblock {\em Kamus Indonesia-Jawa}.
\newblock PT Gramedia Pustaka Utama, Jakarta.

\bibitem[Viberg, 1983]{viberg1983}
Viberg, {\AA}. (1983).
\newblock The verbs of perception: A typological study.
\newblock {\em Linguistics}, 21(1):123--162.

\bibitem[W{\"a}lchli and Cysouw, 2012]{walchli2012}
W{\"a}lchli, B. and Cysouw, M. (2012).
\newblock Lexical typology through similarity semantics: Toward a semantic map of motion verbs.
\newblock {\em Linguistics}, 50(3):671--710.

\bibitem[Wierzbicka, 2007]{wierzbicka2007}
Wierzbicka, A. (2007).
\newblock Bodies and their parts: An {NSM} approach to semantic typology.
\newblock {\em Language Sciences}, 29(1):14--65.

\bibitem[Zaidan and Callison-Burch, 2014]{2014arabic}
Zaidan, O.~F. and Callison-Burch, C. (2014).
\newblock Arabic dialect identification.
\newblock {\em Computational Linguistics}, 40(1):171--202.

\end{thebibliography}





% \end{thebibliography}

\end{document}